\documentclass{article}

\usepackage[final, nonatbib]{neurips_2024}

\usepackage[utf8]{inputenc} 
\usepackage[T1]{fontenc}    
\usepackage[hidelinks]{hyperref}       
\usepackage{url}            
\usepackage{booktabs}       
\usepackage{amsfonts}       
\usepackage{nicefrac}       
\usepackage{microtype}      
\usepackage{xcolor}         
\usepackage[numbers]{natbib}

\usepackage{amsmath,amsfonts,bm}

\def\eqref#1{equation~\ref{#1}}

\def\1{\bm{1}}

\DeclareMathAlphabet{\mathsfit}{\encodingdefault}{\sfdefault}{m}{sl}
\SetMathAlphabet{\mathsfit}{bold}{\encodingdefault}{\sfdefault}{bx}{n}

\usepackage{comment}
\usepackage{xspace}
\usepackage{xcolor}
\usepackage{booktabs}

\usepackage{graphicx}
\usepackage[center]{subfigure}
\usepackage{balance}  

\usepackage{graphics}
\usepackage{xspace}

\usepackage{booktabs}
\usepackage{soul}
\usepackage{balance}

\usepackage{float}
\usepackage{algorithm}
\usepackage[noend]{algpseudocode}
\usepackage{amsmath}
\usepackage{amssymb}
\usepackage{mathtools}
\usepackage{amsfonts}
\usepackage{amsbsy}
\usepackage{amsthm}
\usepackage[]{caption}
\usepackage{multirow}
\usepackage[mathscr]{eucal} 
\usepackage{mathtools}
\usepackage{verbatim}
\usepackage{tabularx}
\usepackage{enumitem}
\usepackage{array}
\usepackage{soul}
\usepackage{dsfont}
\usepackage{url}
\usepackage{bbm}

\usepackage[most]{tcolorbox}

\usepackage{pifont}
\usepackage{xcolor,colortbl}
\usepackage[inkscapeformat=png]{svg}
\usepackage{makecell}
\usepackage{xspace}
\usepackage{footmisc} 
\usepackage{wrapfig}
\hypersetup{
    colorlinks=true,
    linkcolor=black,
    citecolor = black,
    filecolor=magenta,      
    urlcolor=magenta,
    pdftitle={Overleaf Example},
    pdfpagemode=FullScreen,
    }

\definecolor{googleblue}{RGB}{66,133,244}
\definecolor{googlered}{RGB}{219,68,55}
\definecolor{googleyellow}{RGB}{244,180,0}
\definecolor{googlegreen}{RGB}{15,157,88}

\newtcolorbox{mybox}[2][]{%
  attach boxed title to top left
               = {yshift=-8pt},
  colback      = cyan!5!white,
  colframe     = googleblue,
  fonttitle    = \bfseries,
  colbacktitle = googleblue,
  title        = #2,#1,
  enhanced,
}

\newcommand{\method}{\textsc{MuSE}\xspace}

\newcommand{\std}{\scriptsize}

\newtheorem{theorem}{\textbf{Theorem}}

\makeatletter
\newtheorem*{rep@theorem}{\rep@title}
\newcommand{\newreptheorem}[2]{%
\newenvironment{rep#1}[1]{%
 \def\rep@title{#2 \ref{##1}}%
 \begin{rep@theorem}}%
 {\end{rep@theorem}}}
\makeatother

\newreptheorem{theorem}{Theorem}

\definecolor{sunwoogreen}{rgb}{0.66, 0.89, 0.63}
\definecolor{sunwoogreen2}{RGB}{67, 148, 58}
\definecolor{sunwooyellow}{rgb}{1.0, 1.0, 0.0}
\definecolor{sunwooyellow2}{RGB}{228, 208, 10}

\newcommand{\best}{\cellcolor{sunwoogreen}}  
\newcommand{\secb}{\cellcolor{sunwooyellow}}  

\newcommand{\gaemodel}{GAE\xspace}

\AtBeginDocument{%
  \providecommand\BibTeX{{%
    \normalfont B\kern-0.5em{\scshape i\kern-0.25em b}\kern-0.8em\TeX}}}

\title{Rethinking Reconstruction-based Graph-Level Anomaly Detection: Limitations and a Simple Remedy}

\author{%
  Sunwoo Kim$^{1}$, Soo Yong Lee$^{1}$, Fanchen Bu$^{2}$, Shinhwan Kang$^{1}$, \\
  \textbf{Kyungho Kim$^{1}$, Jaemin Yoo$^{2}$, Kijung Shin$^{1,2}$}\\
  $^{1}$Kim Jaechul Graduate School of AI, $^{2}$School of Electrical Engineering\\
  Korea Advanced Institute of Science and Technology (KAIST)\\
  \texttt{\{kswoo97, syleetolow, boqvezen97, shinhwan.kang,} \\ \texttt{kkyungho, jaemin, kijungs\}@kaist.ac.kr} \\
}

\begin{document}

\maketitle

\begin{abstract}
  Graph autoencoders (Graph-AEs) learn representations of given graphs by aiming to accurately reconstruct them.
A notable application of Graph-AEs is graph-level anomaly detection (GLAD), whose objective is to identify graphs with anomalous topological structures and/or node features compared to the majority of the graph population.
Graph-AEs for GLAD regard a graph with a high mean reconstruction error (i.e. mean of errors from all node pairs and/or nodes) as anomalies. 
Namely, the methods rest on the assumption that they would better reconstruct graphs with similar characteristics to the majority.
We, however, report non-trivial counter-examples, a phenomenon we call \textit{reconstruction flip}, and highlight the limitations of the existing Graph-AE-based GLAD methods.
Specifically, we empirically and theoretically investigate when this assumption holds and when it fails.
Through our analyses, we further argue that, while the reconstruction errors for a given graph are effective features for GLAD, leveraging the multifaceted summaries of the reconstruction errors, beyond just mean, can further strengthen the features.
Thus, we propose a novel and simple GLAD method, named \textsc{MuSE}.
The key innovation of \textsc{MuSE} involves taking multifaceted summaries of reconstruction errors as graph features for GLAD.
This surprisingly simple method obtains SOTA performance in GLAD, 
performing best overall among 14 methods across 10 datasets. 
\end{abstract}

\vspace{-4mm}
\section{Introduction}
\vspace{-1mm}
\label{sec:introduction}
Graph autoencoders (Graph-AEs) are a family of graph neural networks (GNNs) that learn latent representations of given graphs by aiming to accurately reconstruct them.
Representative examples of Graph-AEs include \gaemodel~\citep{kipf2016variational}
and GraphMAE~\citep{hou2022graphmae}, which respectively aim to accurately reconstruct graph structure and node features.
Graph-AEs have a broad spectrum of applications, such as anomaly detection~\citep{niu2023graph, luo2022deep, ding2019deep} and link prediction~\citep{guo2022multi, kipf2016variational, kollias2022directed}.

\begin{figure*}[t]
    \subfigure[Training graphs]{\includegraphics[width=0.67\linewidth]{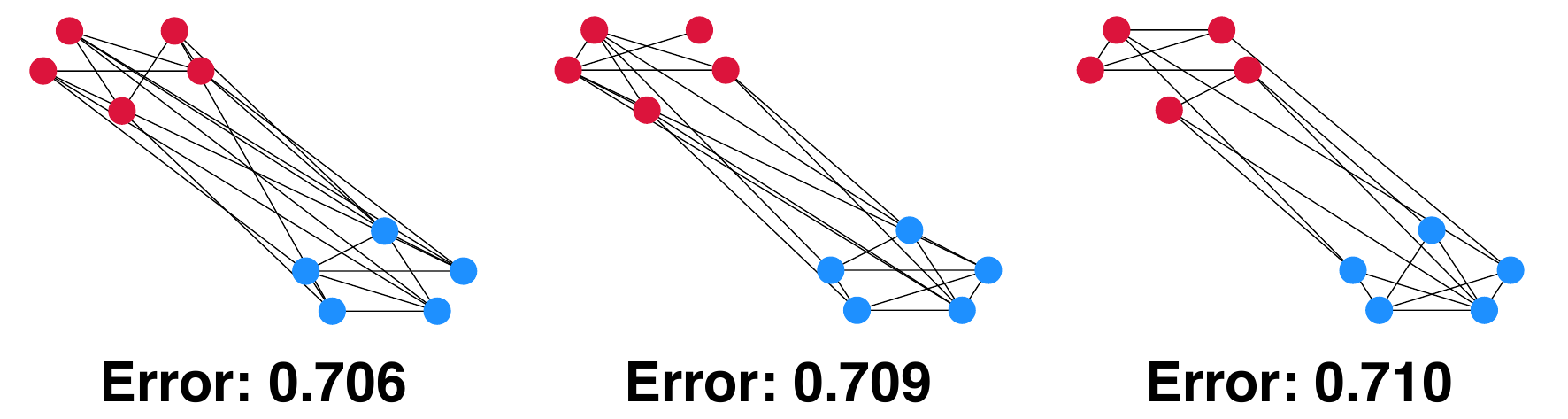}}
    \hspace{12mm}
    \subfigure[Unseen graph]{\includegraphics[width=0.22\linewidth]{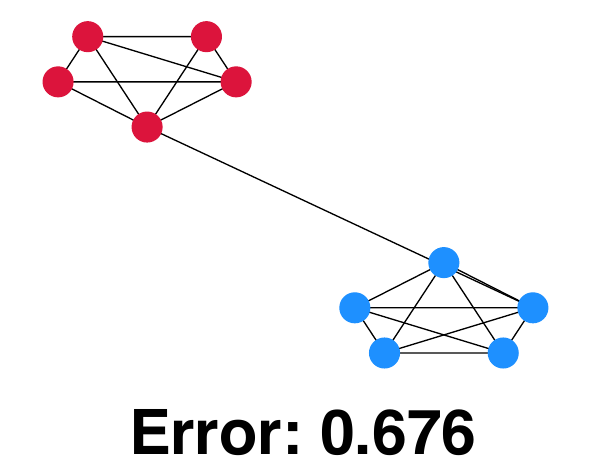}}
    \caption{
    The training graphs in (a) and the unseen graph in (b) exhibit different structural characteristics,
    but a Graph-AE model reconstructs the graph in (b) more accurately than those in (a). 
    }
    \label{fig:introexample}
\end{figure*}

Graph-AEs are trained to capture the structural information of graphs used for training (i.e., training graphs). 
Thus, intuitively, Graph-AEs are expected to better reconstruct graphs that are similar to the training graphs.
However, to our surprise, this expectation is not always true.
Given Graph-AEs trained to reconstruct the graphs in Figure~\ref{fig:introexample}(a), which share common structural characteristics,
one would expect that the Graph-AEs would reconstruct the training graphs with smaller reconstruction errors than a dissimilar counterpart (e.g., in Figure~\ref{fig:introexample} (b)).
We, however, report the opposite.
The reconstruction error is lower for the graph in Figure~\ref{fig:introexample}(b) than the training graphs.
While similar observations were reported in computer vision application~\citep{liu2023diversity}, which we elaborate on in Section~\ref{subsec:relatedwork}, the existing literature does not clearly explain the aforementioned phenomenon.

This counterintuitive phenomenon guides our analysis, implication, and the proposed method.
First, we aim to understand this counterintuitive phenomenon, which we refer to as the \textit{reconstruction flip}.
We argue that the phenomenon can occur when (1) the training graphs and unseen graphs share a primary structural pattern and (2) the pattern is more pronounced in unseen graphs than in training graphs.
We corroborate our claim with our theoretical and empirical analysis.

Second, our finding has strong implications for graph-level anomaly detection (GLAD).
GLAD aims to identify graphs of different {characteristics} (e.g., structures and/or features) from the majority~\citep{niu2023graph, luo2022deep, qiu2022raising, liu2024towards}. 
{Many existing GLAD methods leverage Graph-AEs~\citep{luo2022deep, niu2023graph}.}
They regard a graph with a high {mean} reconstruction error (i.e. mean of errors from all node pairs and/or nodes) as anomalies, assuming trained Graph-AEs would struggle to reconstruct the graphs with structural patterns (and/or node attributes) that deviate from most training graphs.
However, our finding suggests that this assumption does not always hold.
Specifically, anomalous graphs with structural characteristics distinct from the training graphs may exhibit similar or even lower mean reconstruction errors.
In such cases, the existing methods would struggle to detect anomalous graphs.
We argue that this issue arises from how the existing methods leverage reconstruction errors, yet these errors remain valuable graph features for GLAD.
We propose using multifaceted summaries of the reconstruction errors, rather than relying solely on their mean, to further enhance the features.

Third, based on the analysis and implication, we propose a simple and novel GLAD method, \textbf{\method} (\textbf{\underline{Mu}}ltifacted \textbf{\underline{S}}ummarization of Reconstruction \textbf{\underline{E}}rrors). 
\method, like other Graph-AE-based GLAD methods, reconstructs given graphs.
{However, \method employs a new way of using reconstruction errors: representing a graph with multifaceted summaries (e.g., mean, standard deviation, etc.) of the graph's reconstruction errors for GLAD.}
This simple innovation allows \method to obtain SOTA performance in the GLAD tasks.
Through our extensive experiments including 13 baseline methods and 10 benchmark datasets, \method performs overall best, achieving up to 28.1\% performance gain (in terms of AUROC) compared to the best competitor.
We summarize our contribution as follows:
\begin{itemize}[leftmargin = *]
    \item \textbf{Analysis (Section~\ref{sec:analysis}) with implications (Section~\ref{sec:implication}):} {We investigate the reconstruction flip phenomenon both theoretically and empirically, yielding practical implications for GLAD.}
    \item \textbf{Effective method (Section~\ref{sec:method}):} Motivated by our analysis, we propose \method, a simple yet effective GLAD method that represents a graph as multifaceted summaries of its reconstruction errors.
    \item \textbf{Extensive experiments (Section~\ref{sec:experiment}):} Our experiments on $10$ datasets demonstrate the superiority of \method over prior GLAD methods. 
    Our code and datasets are available at~\url{https://github.com/kswoo97/GLAD_MUSE}.
\end{itemize}

\section{Related Work and Preliminaries}
\label{sec:relatedwork}

In this section, we provide a brief literature review on graph-level anomaly detection (GLAD) and peculiar observations in GLAD.
We then present the preliminaries of our work. 

\subsection{Related work}\label{subsec:relatedwork}

\textbf{Graph-level anomaly detection.}
Graph-level anomaly detection (GLAD) aims to identify graphs {whose characteristics deviate from those of the majority of the graph population}~\citep{niu2023graph, luo2022deep, qiu2022raising, liu2024towards}. 
GLAD has various applications, including brain diagnosis~\citep{lanciano2020explainable} and drug discovery~\citep{vogel2002drug}.
While several GLAD methods require anomaly label supervision to train a detector~\citep{zhang2022dual, ma2023towards}, we focus on the methods that do not require anomaly labels~\citep{niu2023graph, luo2022deep, qiu2022raising, liu2024towards}, since anomaly labels are often not available in real-world scenarios.
A GLAD method typically trains a model to perform certain pretext tasks on given graphs, 
and the graphs are regarded as anomalies if the trained model shows poor pretext task performance on them.
Notable pretext tasks include graph reconstruction~\citep{niu2023graph, luo2022deep}, graph embedding hypersphere minimization~\citep{qiu2022raising, zhao2022graph}, and cross-view mutual information maximization~\citep{ma2022deep, liu2024towards}.



\textbf{Peculiar observations in graph-level anomaly detection.}
The most relevant literature with our work is a study of~\citet{zhao2023using}.
In graph classification datasets, they regarded graphs belonging to a particular class as normal graphs and otherwise as anomalies to benchmark several GLAD methods.
{In this setting, they observed that several GLAD methods, especially kernel-based ones, occasionally perform worse than random guessing.}
However, their analysis focused on graph kernels~\citep{neumann2016propagation, shervashidze2011weisfeiler}, without covering graph autoencoders, which are our focus.
In computer vision,~\citet{liu2023diversity} suggested that certain anomalous images can be easier to reconstruct than normal ones, presenting a method that can overcome these undesirable circumstances.
However, neither their analysis nor the method can be trivially extended to the graph domain, as elaborated in Appendix~\ref{subsec:cvreconflip}.

\subsection{Preliminary}\label{subsec:prelim}

\textbf{Graphs.} 
A graph $\mathcal{G} = (\mathcal{V},\mathcal{E})$ is defined by a node set $\mathcal{V} = \{v_{1},\ldots,v_{\vert \mathcal{V} \vert}\}$ and an edge set $\mathcal{E} = \{e_{1},\ldots ,e_{\vert \mathcal{E} \vert}\}$, where each edge $e_{i} \in \mathcal{E}$ is defined by a node pair.
We assume that each node $v_{i} \in \mathcal{V}$ is equipped with a feature vector $\mathbf{x}_{i} \in \mathbb{R}^{d}$,
which forms a node feature matrix $\mathbf{X} \in \mathbb{R}^{\vert \mathcal{V} \vert \times d}$ where each $i$-th row $\mathbf{X}_{i,:} = \mathbf{x}_{i}$.
Moreover, $\mathcal{E}$ can be represented by an adjacency matrix $\mathbf{A}\in \{0,1\}^{\vert \mathcal{V}\vert \times \vert \mathcal{V}\vert}$, where $\mathbf{A}_{i,j}=1$ if and only if $\{v_{i},v_{j}\} \in \mathcal{E}$.
{Therefore, the graph can also be defined as $\mathcal{G} = (\mathbf{X}, \mathbf{A})$.}

\textbf{Graph neural networks.} 
Graph neural networks (GNNs) are a category of neural networks designed for processing graph data. 
GNNs primarily leverage message passing schemes~\citep{xu2018powerful, hamilton2017inductive, velivckovic2017graph, gilmer2017neural, lee2023towards, liang2024sign}. 
In this work, we consider GNNs as parameterized functions $f_{\theta}$ that generate node embeddings $\mathbf{Z} = f_{\theta}(\mathbf{X}, \mathbf{A}) \in \mathbb{R}^{\vert \mathcal{V}\vert \times d'}$ for each graph $\mathcal{G} = (\mathbf{X},\mathbf{A})$.~\footnote{Some GNNs are designed to produce edge embeddings~\citep{zhou2023co, zhou2023co} and/or graph embeddings~\citep{ying2018hierarchical, gao2019graph}.}

\textbf{Graph autoencoders.}
Graph autoencoders (Graph-AEs) are a family of GNNs that learn graph latent representations by graph reconstruction.
Specifically, Graph-AEs reconstruct a given graph's structure~\cite{kipf2016variational, tan2023s2gae} and/or node features~\cite{hou2022graphmae, hou2023graphmae2}.
For example, \gaemodel~\citep{kipf2016variational} reconstructs the adjacency matrix of a given graph.
Specifically, \gaemodel first uses an encoder GNN $f_{\theta}$ to
generate node embeddings $\mathbf{Z} = f_{\theta} (\mathbf{X}, \mathbf{A})$ of a given graph $\mathcal{G} = (\mathbf{X}, \mathbf{A})$.
Then, \gaemodel obtains a reconstructed adjacency matrix $\hat{\mathbf{A}} \in \mathbb{R}^{\vert \mathcal{V}\vert \times \vert \mathcal{V} \vert}$ through the inner-product of node embeddings and entry-wise sigmoid activation $\sigma$ (i.e., $\hat{\mathbf{A}} = \sigma(\mathbf{Z}\mathbf{Z}^{T})$). 
Lastly, it minimizes the difference between $\mathbf{A}$ and $\hat{\mathbf{A}}$, for which one can use the binary cross-entropy (BCE) loss 
$\mathcal{L}_{\mathrm{BCE}}(\mathcal{G}~\vert~f_{\theta}) \coloneqq -\sum_{i,j =1}^{\vert \mathcal{V} \vert}\left(\mathbf{A}_{i,j}\log{\hat{\mathbf{A}}_{i,j}} + (1 - \mathbf{A}_{i,j})\log{(1-\hat{\mathbf{A}}_{i,j})}\right)$ or the squared Frobenius-norm loss $\mathcal{L}_{\mathrm{SFN}}(\mathcal{G}~\vert~f_{\theta}) \coloneqq \lVert \mathbf{A} - \hat{\mathbf{A}} \rVert^2_{F} = \sum_{i,j =1}^{\vert \mathcal{V} \vert}\left({\mathbf{A}}_{i,j}-\hat{\mathbf{A}}_{i,j}\right)^2$.
Given a set of training graphs $\mathbb{G} = \{\mathcal{G}_{1},\mathcal{G}_{2},\cdots,\mathcal{G}_{\vert \mathbb{G}\vert} \}$, \gaemodel is trained (i.e., the encoder parameter $\theta$ is updated) to minimize $\sum_{\mathcal{G} \in \mathbb{G}} \mathcal{L}(\mathcal{G}~\vert~f_{\theta})$ (or equivalently, $\mathbb{E}_{\mathcal{G} \in \mathbb{G}} \mathcal{L}(\mathcal{G}~\vert~f_{\theta})$), where $\mathcal{L}$ is $\mathcal{L}_{\mathrm{BCE}}$ or $\mathcal{L}_{\mathrm{SFN}}$.

In our analysis below, we focus on methods that reconstruct graph structures, while we describe and investigate node-feature-reconstruction methods in Appendix~\ref{subsec:nodefeat}.

\section{Analysis of Graph Autoencoders}
\label{sec:analysis}

In this section, we explore \textit{reconstruction flip} phenomena (e.g., the phenomenon described in Figure~\ref{fig:introexample}),
where graph autoencoders (Graph-AEs) reconstruct some unseen graphs {that are dissimilar} to training graphs better than training graphs.

Our analysis focuses on evidencing the claims below. 
We assume \gaemodel~\cite{kipf2016variational} trained on graphs sharing a primary pattern $\mathcal{P}$ of strength $\mathcal{S}$.
\begin{mybox}[colback=googleblue!10!white,colframe=googleblue]{Claims}\label{box:claim}
  \begin{itemize}[leftmargin=*]
      \item \textit{Reconstruction flip \textbf{tends to occur} when unseen (test) graphs have the same primary pattern $\mathcal{P}$ but with a greater strength $\mathcal{S}' > \mathcal{S}$.}
      \item \textit{Reconstruction flip \textbf{tends not to occur} when unseen (test) graphs have a different primary pattern $\mathcal{P}' \neq \mathcal{P}$.}
  \end{itemize}
\end{mybox}

That is, when Graph-AEs are trained on graphs with a primary pattern $\mathcal{P}$ of strength $\mathcal{S}$, the trained Graph-AEs tend to exhibit lower errors in reconstructing graphs with $\mathcal{P}$ of a greater strength $\mathcal{S}' > \mathcal{S}$.
However, the trained Graph-AEs tend to exhibit higher errors in reconstructing graphs with a different pattern $\mathcal{P}' \neq \mathcal{P}$.
For demonstration, we elaborate on $\mathcal{P}$ and $\mathcal{S}$ via synthetic datasets (Section~\ref{subsec:analysissetting}) and present both empirical (Section~\ref{subsec:empiricalanalysis}) and theoretical (Section~\ref{subsec:theoreticalanalysis}) investigations.

\subsection{Synthetic graphs}\label{subsec:analysissetting}

To elaborate on the concepts of a primary pattern $\mathcal{P}$ and a pattern strength $\mathcal{S}$, 
we employ two types of synthetic graphs: (1) graphs with community structures and (2) graphs with cycles.

\textbf{Community type.}
Syn-Com graphs have community structures, {a pervasive pattern in real-world graphs}~\citep{girvan2002community}, as the primary pattern $\mathcal{P}$.
Graphs in Figure~\ref{fig:introexample} are instances of such graphs.
We control the strength $S$ of community structures through a parameter $\tau \in [0, 1]$.
The $10$ nodes in each graph are evenly divided into two communities.
Each intra- and inter-community edge is formed with probability $(1 + \tau)/2$ and $(1 - \tau)/2$, respectively.
In our empirical analysis, we consider two graph classes with different $\tau$'s. 
Graphs in one class $\mathbb{G}^{com}_{\tau = 0.4}$ are {generated} with $\tau=0.4$ (weaker community structure; the graphs in Figure~\ref{fig:introexample}(a)) and those in the other class $\mathbb{G}^{com}_{\tau = 0.8}$ are {generated} with $\tau=0.8$ (stronger community structure; the graph in Figure~\ref{fig:introexample}(b)).

\begin{figure}[t] 
    \centering
    \includegraphics[width=\linewidth]{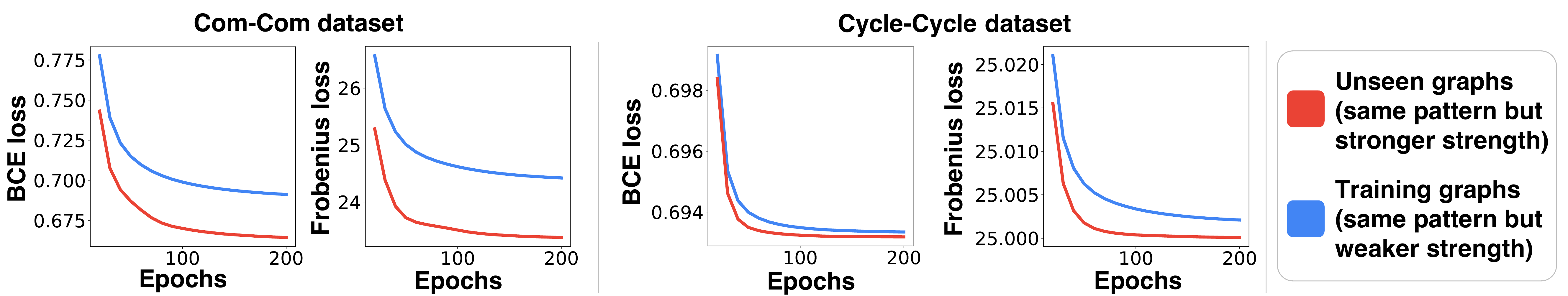}
    \caption{\label{fig:empiricallyreal}
    \textbf{Reconstruction flip occurs.}
    When Graph-AEs are trained on graphs sharing a primary pattern of weak strength, the trained Graph-AEs exhibit lower reconstruction errors for graphs with the same pattern but with higher strength (\textcolor{googlered}{red} lines) than those with weaker strength (\textcolor{googleblue}{blue} lines).
    }
\end{figure}

\begin{figure}[t] 
    \centering
    \includegraphics[width=\linewidth]{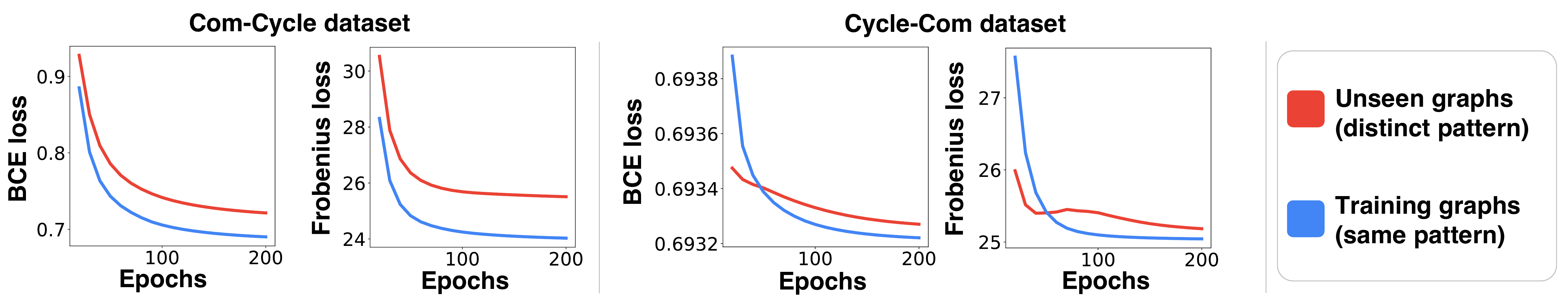}
    \caption{\label{fig:empiricallyfine}
    \textbf{Reconstruction flip does NOT occur.}
    When Graph-AEs are trained on graphs sharing a primary pattern, the trained Graph-AEs exhibit higher reconstruction errors for graphs with a different pattern (\textcolor{googlered}{red} lines) than those with the same pattern (\textcolor{googleblue}{blue} lines).}
\end{figure}

\begin{wrapfigure}{r}{0.35\linewidth} 
    \centering
    \vspace{-5mm}
    \includegraphics[width=\linewidth]{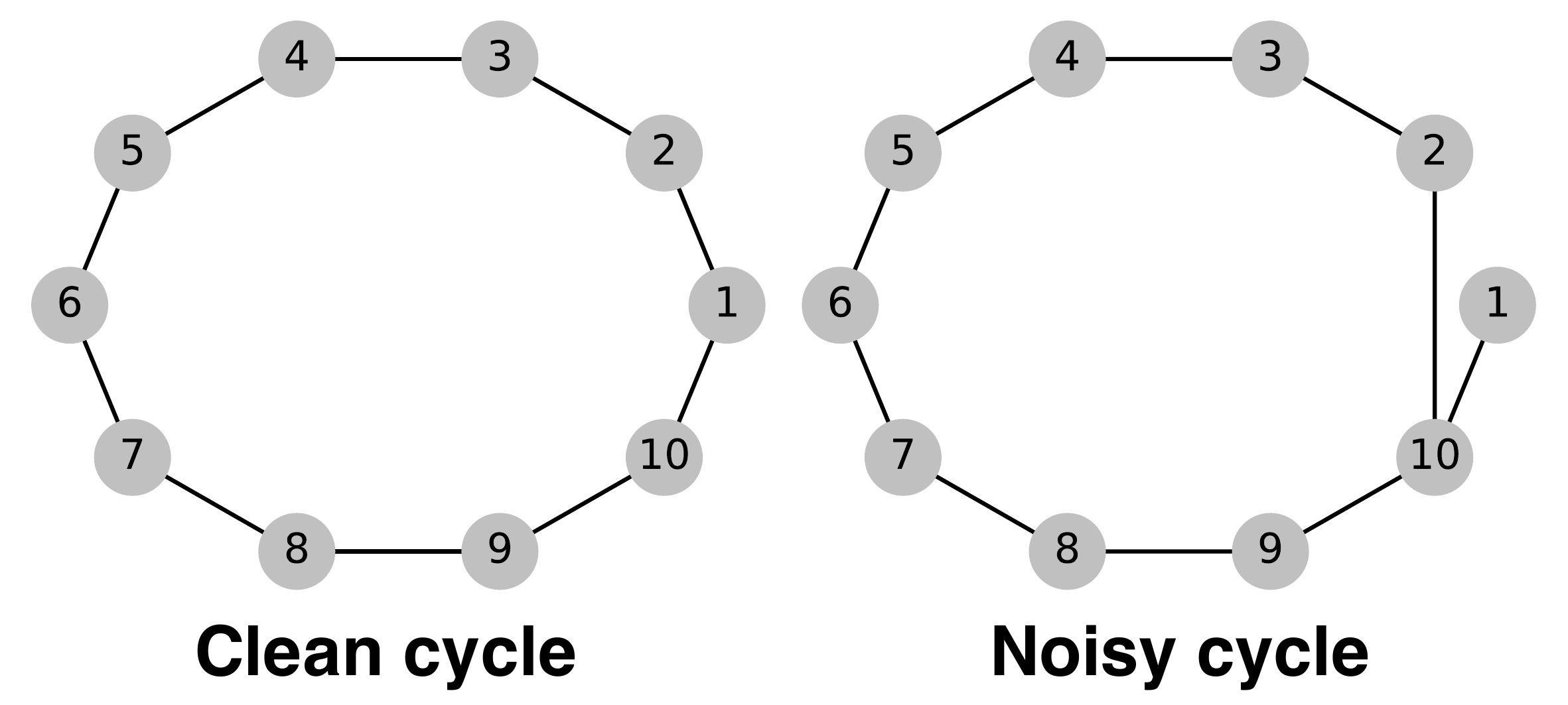}
    \caption{\label{fig:motif}A clean-cycle graph and a noisy-cycle graph.}
    \vspace{-5mm}
\end{wrapfigure}

\textbf{Cycle type.} 
Syn-Cycle graphs contain a cycle of nodes (i.e., a closed chain of nodes), which is commonly observed in real-world graphs~\cite{farkas2001spectra}.
In Syn-Cycle, the primary pattern $\mathcal{P}$ corresponds to the node cycle.
Graphs in Figure~\ref{fig:motif} are instances of Syn-Cycle.
In Syn-Cycle graphs, the pattern strength $\mathcal{S}$ is stronger in the clean-cycle graph consisting only of a cycle with $n$ nodes (the left graph in Figure~\ref{fig:motif}),
and it is weaker in noisy-cycle graphs, where a node within the cycle of ($n-1$) nodes is linked to an extra node (the right graph in Figure~\ref{fig:motif}).
The number of nodes $n$ is fixed to $10$.

We provide detailed descriptions of both graph types in Appendix~\ref{sec:syntheticformal}.

\subsection{Empirical analysis}\label{subsec:empiricalanalysis}

\textbf{Setting.} 
We consider two scenarios.
In Scenario 1, we aim to validate our first claim.
To this end, we {leverage} (1) the Com-Com dataset, where the training graphs are sampled from $\mathbb{G}^{com}_{\tau = 0.4}$ and the unseen graphs are sampled from $\mathbb{G}^{com}_{\tau = 0.8}$, and (2) the Cycle-Cycle dataset, where the training graphs are sampled from the noisy-cycle graphs and the unseen graph is the clean-cycle graph.

In Scenario 2, we aim to validate our second claim.
To this end, we {leverage} (3) the Com-Cycle dataset, where the training graphs are sampled from $\mathbb{G}^{com}_{\tau = 0.4}$ and the unseen graphs are sampled from the noisy-cycle graphs, and (4) the Cycle-Com dataset, where the training graphs are sampled from the noisy-cycle graphs and the unseen graphs are sampled from $\mathbb{G}^{com}_{\tau = 0.4}$.

For each dataset, we train GAE~\citep{kipf2016variational} equipped with GIN~\citep{xu2018powerful} as the {graph} encoder. 
We use either the (1) binary cross-entropy (BCE) loss or the (2) squared Frobenius-norm (Frobenius) loss, which are formalized in Section~\ref{subsec:prelim}, for the reconstruction loss. 
We compare the reconstruction losses for the training graphs and unseen graphs.
Further details are provided in Appendix~\ref{subsec:analysisdetails}.

\textbf{\textit{Note.}} \textit{While we focus on adjacency reconstruction in this section, we provide an analysis of feature-reconstruction methods in Appendix~\ref{subsec:nodefeat}, where the results are consistent with those below.}

\textbf{Empirical results.}
In {Scenario 1}, for both Com-Com and Cycle-Cycle datasets, reconstruction losses are lower for the graphs with stronger pattern strengths, which are unseen graphs (see Figure~\ref{fig:empiricallyreal}).
That is, all graphs share a primary pattern $\mathcal{P}$ (either node community or cycle) with two different strengths $\mathcal{S} < \mathcal{S}'$, and \gaemodel trained on graphs of pattern strength $\mathcal{S}$ reconstructs those of pattern strength $\mathcal{S}'$ with lower losses, demonstrating our first claim.
In {Scenario 2}, for both Com-Cycle and Cycle-Com datasets, reconstruction losses become lower for the training graphs after some training epochs (see Figure~\ref{fig:empiricallyfine}).
That is, given training graphs with a primary pattern $\mathcal{P}$ (either node communities or a cycle), the trained \gaemodel reconstructs graphs with the same pattern $\mathcal{P}$ with lower losses than those with a different pattern $\mathcal{P}'$, demonstrating our second claim.

\subsection{Theoretical analysis}\label{subsec:theoreticalanalysis}

In this subsection, we theoretically analyze the empirical observations in Section~\ref{subsec:empiricalanalysis}, focusing on the occurrence of reconstruction flip.
Informally speaking, we shall show that when \gaemodel is trained on graphs with a primary pattern $\mathcal{P}$ of strength $\mathcal{S}$, the following holds:
\begin{enumerate}[leftmargin=*,label={(\bfseries A\arabic*)}] 
    \item Reconstruction losses decrease for graphs with the same $\mathcal{P}$ of various strengths.\label{item:generalization}
    \item Reconstruction losses decrease more rapidly for graphs with the same $\mathcal{P}$ of greater strengths.\label{item:decrease}
\end{enumerate}

\textbf{Setting.}
For theoretical analysis, we use a single-layer linear GAE.
Formally, for a given graph $\mathcal{G} = (\mathbf{X}, \mathbf{A})$, the linear GAE reconstructs the given graph's adjacency matrix as follows: $\hat{\mathbf{A}} = \mathbf{AXW}(\mathbf{AXW})^{T}$, where $\mathbf{W} \in \mathbb{R}^{n \times n}$ is a learnable weight matrix and $n$ is the number of nodes.
We use the squared Frobenius norm as the reconstruction loss $\mathcal{L}(\mathcal{G}, \mathbf{W}) = \lVert \mathbf{A} - \hat{\mathbf{A}}\rVert^{2}_{F}$.
We consider Syn-Com graphs $\mathbb{G}^{com}_{\tau}$ (Section~\ref{subsec:analysissetting}).
To formalize the training of linear GAE, we define the expected gradient descent update of $\mathbf{W}$ as follows: 
\begin{equation}\label{eq:parameterupdate}
    \mathbb{W}(\tau, \mathbf{W}, \gamma) = \mathbf{W} -\gamma \mathbb{E}_{\mathcal{G}}\left[\frac{\partial \mathcal{L}(\mathcal{G}, \mathbf{W})}{\partial \mathbf{W}} \ \vert \ \tau\right] \in \mathbb{R}^{n \times n},    
\end{equation}
where $\gamma > 0$ is a learning rate and $\mathbb{E}_{\mathcal{G}}[\cdot \vert \tau]$ takes the expectation over all the graphs $\mathcal{G} \in \mathbb{G}^{com}_{\tau}$.
We assume graphs of strength $\tau_1$ are used as the training graphs and
graphs of strength $\tau_2$ are used as the unseen (test) graphs.

\textbf{Theoretical results.}
We now present our theoretical results, with proof of each theorem in Appendix~\ref{sec:appendix_proof}.
We first demonstrate that \ref{item:generalization} holds by expectation.
\begin{theorem}[Generalization across $\tau$]\label{theorem:generalization}
    For any $n \geq 12$ and $0 < \tau_1 \leq 1$, 
    there exists $\epsilon > 0$ such that for any $0 < \gamma \leq \epsilon$ 
    and any $0 < \tau_2 \leq 1$ (recall that $0$ and $1$ are the lower and upper bounds of $\tau$, respectively)
    with the initial weight matrix $\mathbf{W}^{(0)} = \mathbb{I}_{n}$,
    \begin{equation*}
        \underbrace{\mathbb{E}_{\mathcal{G}} [\mathcal{L}(\mathcal{G},\mathbb{W}(\tau_{1}, \mathbf{W}^{(0)}, \gamma)) \vert \tau_{2}]}_{\text{the loss on graphs of strength $\tau_2$ \textbf{after} update}} < 
        \underbrace{\mathbb{E}_{\mathcal{G}} [\mathcal{L}(\mathcal{G},\mathbf{W}^{(0)}) \vert \tau_{2}]}_{\text{the loss on graphs of strength $\tau_2$ \textbf{before} update}}.
    \end{equation*}
\end{theorem}

Roughly, Theorem~\ref{theorem:generalization} states that regardless of the values of $\tau_1$ and $\tau_2$, updating $\mathbf{W}$ with graphs of strength $\tau_1$ reduces the reconstruction losses for graphs of strength $\tau_2$.
Intuitively, this suggests that GAE learns the primary pattern $\mathcal{P}$, regardless of the strength $\mathcal{S}$ (here, $\tau_1$) of the training graphs.


Now, we demonstrate that \ref{item:decrease} holds by expectation.
\begin{theorem}[Correlation with $\tau$]\label{theorem:decrease}
    For any $n \geq 34$ and $0 < \tau_1 \leq 0.98$, there exists $\epsilon > 0$
    such that for any $0 < \gamma \leq \epsilon$ and any $\tau_1 + 0.02 \leq \tau_2 \leq 1$ with the initial weight $\mathbf{W}^{(0)} = \mathbb{I}_{n}$,
    \begin{equation*}
       \underbrace{\mathbb{E}_{\mathcal{G}} \left[\mathcal{L}(\mathcal{G},\mathbf{W}^{(0)}) -  \mathcal{L}(\mathcal{G},\mathbb{W}(\tau_{1}, \mathbf{W}^{(0)},\gamma))~\vert~\tau_{1}\right]}_{\text{the decrease in loss on graphs of strength $\tau_1$ (training graphs)}} <  \underbrace{\mathbb{E}_{\mathcal{G}}\left[\mathcal{L}(\mathcal{G},\mathbf{W}^{(0)}) -  \mathcal{L}(\mathcal{G},\mathbb{W}(\tau_{1}, \mathbf{W}^{(0)},\gamma))~\vert~\tau_{2}\right]}_{\text{the decrease in loss on graphs of strength $\tau_2$ (unseen graphs)}}.
    \end{equation*}
\end{theorem}

Roughly, Theorem~\ref{theorem:decrease} states that updating $\mathbf{W}$ using graphs of pattern strength $\tau_1$ results in a greater reduction of reconstruction losses for graphs of higher strength $\tau_2 > \tau_1$ than for the graphs used in $\mathbf{W}$ update (i.e., graphs of strength $\tau_{1}$).
Intuitively, this result suggests that even if GAE is trained on graphs with a primary pattern $\mathcal{P}$ of strength $\mathcal{S}$ (here, $\tau_1$),
{the trained GAE tends to reconstruct graphs of a greater strength $\mathcal{S}'$ (here, $\tau_2$) with smaller losses.}

\section{Implications in Graph-Level Anomaly Detection}
\label{sec:implication}
In this section, we investigate the practical implications of our analysis in Section~\ref{sec:analysis} for graph-level anomaly detection (GLAD).
Specifically, we discuss the limitations of existing graph-autoencoder-based GLAD methods and propose an improved approach to using graph reconstruction errors. 

\begin{figure*}[t]
    \vspace{-6mm}
    \subfigure[Graph $\mathcal{G}_{1}$]{\includegraphics[width=0.28\linewidth]{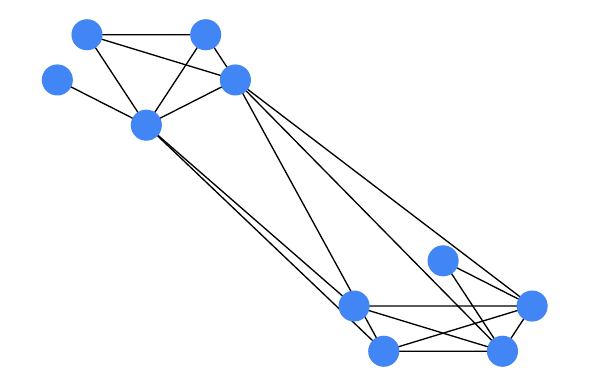}}
    \hspace{2mm}
    \subfigure[Graph $\mathcal{G}_{2}$]{\includegraphics[width=0.28\linewidth]{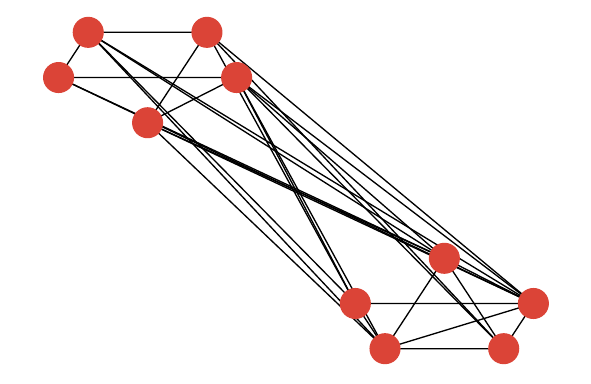}}
    \hspace{5mm}
    \subfigure[Gaussian KDE visualization of error distributions of $\mathcal{G}_{1}$ (\textcolor{googleblue}{blue}) and $\mathcal{G}_{2}$ (\textcolor{googlered}{red}).
    ]{\includegraphics[width=0.38\linewidth]{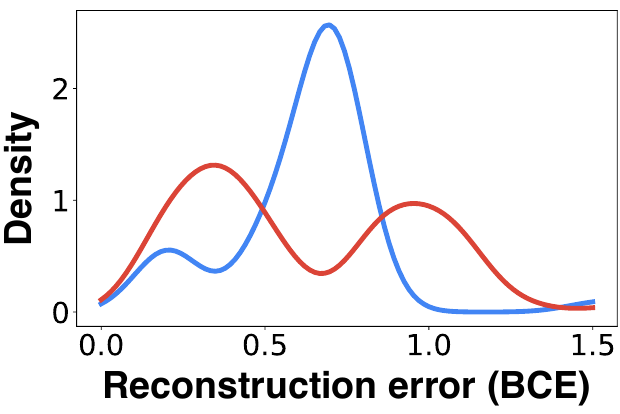}}
    \caption{\label{fig:avgproblem}
    A case of Graph-AEs (specifically, GAE~\citep{kipf2016variational}) having similar mean reconstruction errors for two dissimilar graphs (specifically, \textbf{$\mathcal{G}_{1}$ has 0.6622} and \textbf{$\mathcal{G}_{2}$ has 0.6627}),
    while their error distributions differ significantly.
    }
\end{figure*}

\subsection{Limitations of existing Graph-AEs in GLAD}\label{subsec:baselineimplication}

\textbf{Anomalous graphs may have low mean reconstruction errors.} 
As outlined in Section~\ref{subsec:relatedwork}, many methods for graph-level anomaly detection (GLAD) use graph autoencoders (Graph-AEs), aiming to identify graphs whose characteristics deviate from the majority.
Typically, such GLAD methods~\citep{luo2022deep, niu2023graph} consider graphs with higher {reconstruction losses (i.e. mean (or sum) of reconstruction errors from all node pairs and/or nodes) as anomalies.}
This is based on the intuition that Graph-AEs would struggle to reconstruct anomalous graphs that are dissimilar to the majority of training graphs.
However, our empirical (Section~\ref{subsec:empiricalanalysis}) and theoretical (Section~\ref{subsec:theoreticalanalysis}) analyses reveal that Graph-AEs can exhibit lower mean reconstruction errors for graphs that are less similar to training graphs.
In the context of GLAD, when anomalous graphs have the same primary pattern as training graphs but have stronger strengths, Graph-AEs tend to exhibit lower mean reconstruction errors, compared to the training graphs.
Thus, such anomalies tend not to be detected by existing Graph-AE-based GLAD methods, revealing their severe limitations.

\textbf{Mean is not all you need.} 
In addition in practice, we find that in some cases, dissimilar graphs can have similar mean reconstruction errors.
As shown in Figure~\ref{fig:avgproblem}, although the graph $\mathcal{G}_{1}$ in (a) and the graph $\mathcal{G}_{2}$ in (b) have significantly different numbers of edges, their mean reconstruction errors are similar, making it difficult to distinguish between the two graphs (refer to the caption of Figure~\ref{fig:avgproblem}).
On the other hand, Figure~\ref{fig:avgproblem}(c) shows that the reconstruction error distributions over individual node pairs
\footnote{An example error distribution is the distribution of $\mathbf{A}_{i,j}\log{\hat{\mathbf{A}}_{i,j}} + (1-\mathbf{A}_{i,j})\log{(1-\hat{\mathbf{A}}_{i,j})},\forall i,j\in[n]$.} of $\mathcal{G}_{1}$ (\textcolor{googleblue}{blue distribution}) and $\mathcal{G}_{2}$ (\textcolor{googlered}{red distribution}) differ significantly in shape.
Other descriptive statistics, such as standard deviation, can effectively distinguish the two graphs in such cases.
Thus, incorporating additional statistical measures beyond the mean is {beneficial} for Graph-AEs in GLAD.

\subsection{Enhancing the use of reconstruction errors}\label{subsec:methodmotivation}

Based on the implications discussed in Section~\ref{subsec:baselineimplication}, 
we argue that \textit{a graph's reconstruction errors aggregated in various ways} can provide features that effectively distinguish anomalous graphs from normal graphs.
We provide detailed descriptions of this claim below.

\textbf{Reconstruction errors serve as good features for GLAD.}
Consider the following scenarios where Graph-AEs are trained on graphs sharing a pattern $\mathcal{P}$ of strength $\mathcal{S}$.
First, for an unseen graph of a pattern $\mathcal{P}$ with higher strength $\mathcal{S}' > \mathcal{S}$, Graph-AEs tend to return lower mean reconstruction errors compared to those of the training graphs (Figure~\ref{fig:empiricallyreal}).
Second, for an unseen graph of a different pattern $\mathcal{P}' \neq \mathcal{P}$, Graph-AEs tend to give higher mean reconstruction error than those of the training graphs (Figure~\ref{fig:empiricallyfine}).
In both cases, the error distributions of anomalous graphs differ from those of normal graphs, serving as effective features for distinguishing them.

\textbf{Using multifaceted summaries of errors is important for GLAD.}
In practice, we need a fixed-size representation of the error distribution to be used with anomaly classifiers, which generally require a fixed-size input vector.
To achieve this, we propose using various summary statistics (e.g., mean and standard deviation) to represent the given error distribution.
Specifically, rather than relying on a single statistic, we use multifaceted summaries that capture various aspects of the error distributions.
This is because distinct error distributions may share certain similar summary statistics, as shown in Figure~\ref{fig:avgproblem}.
Therefore, multifaceted summaries of the error distribution can act as an effective feature vector for distinguishing anomalous graphs from normal graphs.


\section{Proposed Method: \method}
\label{sec:method}

Based on the discussions provided in Section~\ref{subsec:methodmotivation}, we present \textbf{\method} (\textbf{\underline{Mu}}ltifacted \textbf{\underline{S}}ummarization of Reconstruction \textbf{\underline{E}}rrors), a simple yet effective graph-level anomaly detection (GLAD) method.
\method aggregates a graph's reconstruction errors into multiple summary statistics and uses a vector of these aggregated errors as the graph's representation for GLAD.
We describe how we train the reconstruction model in Section~\ref{subsec:encodingdecoding} and how we obtain the error representation (i.e., a multifaceted summary of reconstruction errors) in Section~\ref{subsec:errorrepresentation}.


\subsection{Step 1: Training a reconstruction model}
\label{subsec:encodingdecoding}

We first train \method to reconstruct graphs in training data.
Step 1 is composed of three parts: (1) augmentation, (2) encoding/decoding, and (3) reconstruction loss minimization.
For an input graph $\mathcal{G} = (\mathbf{X}, \mathbf{A})$, \method first augments the input graph by randomly dropping some edges, mitigating potential overfitting problems.
Specifically, \method randomly samples $\lceil p \vert \mathcal{E}\vert \rceil$ edges from $\mathcal{E}$ for some $0 \leq p < 1$.
Let $\mathcal{E}'$ denote the set of sampled edges. 
We mask the corresponding entries in $\mathbf{A}$ {by setting them to zero} to generate the augmented adjacency matrix $\mathbf{A}' \in \{0,1\}^{\vert \mathcal{V} \vert \times \vert \mathcal{V} \vert}$ (i.e., $\mathbf{A}'_{i,j} = 0$ if $\{v_{i},v_{j}\} \in \mathcal{E}'$, otherwise $\mathbf{A}'_{i,j} = \mathbf{A}_{i,j, },\forall i,j \in [\vert \mathcal{V}\vert]$).

\method then encodes the augmented graph $\mathcal{G}'=(\mathbf{X}, \mathbf{A}')$ to obtain node embeddings $\mathbf{Z} \in \mathbb{R}^{\vert \mathcal{V} \vert \times d'}$ by using a graph neural network (GNN) encoder $f_{\theta}$ (i.e., $\mathbf{Z} = f_{\theta}(\mathbf{X},\mathbf{A}')$).
Subsequently, \method decodes the node embeddings {$\mathbf{Z}$} to obtain reconstructed node features $\hat{\mathbf{X}} \in \mathbb{R}^{\vert \mathcal{V} \vert \times d}$ and reconstructed adjacency matrix $\hat{\mathbf{A}} \in (0,1)^{\vert \mathcal{V} \vert \times \vert \mathcal{V} \vert}$.
For node feature reconstruction, \method uses a node feature decoder $g_{\psi} : \mathbb{R}^{d'} \mapsto \mathbb{R}^{d}$ (i.e., $\hat{\mathbf{X}} = g_{\psi}(\mathbf{Z})$).
Similarly, for adjacency matrix reconstruction, \method uses an adjacency matrix decoder $h_{\phi} : \mathbb{R}^{d'} \mapsto \mathbb{R}^{d''}$ and the inner product of its output (i.e., $\hat{\mathbf{A}} = \sigma(\mathbf{Z'}\mathbf{Z}^{'T})$, where $\sigma$ is a sigmoid function and $\mathbf{Z'} = h_{\phi}(\mathbf{Z})$).

Lastly, for training, we leverage reconstruction losses regarding the node features and the adjacency matrix.
For node feature reconstruction, we use the cosine-similarity loss as in GraphMAE~\citep{hou2022graphmae}. 
Formally, the loss is defined as follows:
\begin{equation}\label{eq:featmainloss}
    \mathcal{L}_{\mathbf{X}}(\mathcal{G}) = \frac{1}{\vert \mathcal{V}\vert} \sum_{i=1}^{\vert \mathcal{V}\vert} \left(1 - \frac{\mathbf{X}_{i,:}^{T}\hat{\mathbf{X}}_{i,:}}{\lVert \mathbf{X}_{i,:} \rVert_{2} \cdot \lVert \hat{\mathbf{X}} _{i,:} \rVert_{2}}\right), \text{ where $\lVert \cdot \rVert_{2}$ is the $\ell_2$-norm.}
\end{equation}
For adjacency-matrix reconstruction, we use the binary cross-entropy (BCE) loss as in GAE~\citep{kipf2016variational}.
Since real-world graphs are often sparse (i.e., the adjacency matrix contains significantly more zero entries than ones), we weight the loss terms associated with {non-zero entries} to stabilize model training in the presence of this imbalance.
Formally, the loss is defined as follows:
\begin{equation}\label{eq:edgemainloss}
     \mathcal{L}_{\mathbf{A}}(\mathcal{G}) = -\left(\frac{1}{\vert \mathcal{V}\vert}\right)^{2} \sum^{\vert \mathcal{V} \vert}_{i=1}\sum^{\vert \mathcal{V} \vert}_{j=1}\left( \omega\mathbf{A}_{i,j}\log{\hat{\mathbf{A}}_{i,j}} + (1 - \mathbf{A}_{i,j})\log{(1 - \hat{\mathbf{A}}_{i,j})}\right),
\end{equation}
where $\omega = (\frac{\vert \mathcal{V} \vert^{2}}{\sum^{\vert \mathcal{V} \vert}_{i=1}\sum^{\vert \mathcal{V} \vert}_{j=1}\mathbf{A}_{i,j}} - 1)^{\tau}$ and $\tau$ is a scale hyperparameter.
Then, the total reconstruction loss $\mathcal{L}(\mathcal{G})$ is defined as the mean of $\mathcal{L}_{\mathbf{X}}(\mathcal{G})$ (Eq.~(\ref{eq:featmainloss})) and $\mathcal{L}_{\mathbf{A}}(\mathcal{G})$ (Eq.~(\ref{eq:edgemainloss})) (i.e., $\mathcal{L}(\mathcal{G}) = (\mathcal{L}_{\mathbf{X}}(\mathcal{G}) + \mathcal{L}_{\mathbf{A}}(\mathcal{G}))/2$). 
For a set of training graphs $\mathbb{G} = \{\mathcal{G}_{1},\cdots,\mathcal{G}_{K}\}$, we update the parameters (i.e., $\theta, \psi,$ and $\phi$) by using gradient descent aiming to minimize $\frac{1}{K}\sum_{t \in [K]}\mathcal{L}(\mathcal{G}_{t})$.

\subsection{Step 2: Generating error representation}\label{subsec:errorrepresentation}

After training the reconstruction model, we represent each graph using its reconstruction errors, specifically those from all node pairs and/or nodes.
Consider obtaining the reconstruction errors for a graph $\mathcal{G} = (\mathbf{X}, \mathbf{A})$ with the trained reconstruction model (i.e., $f_{\theta},g_{\psi},$ and $h_{\phi}$).
To this end, we first obtain reconstructed features $\hat{\mathbf{X}}$ and a reconstructed adjacency matrix $\hat{\mathbf{A}}$ by using the encoding and decoding scheme described in Section~\ref{subsec:encodingdecoding}.
Note that we do not perform augmentation for Step 2 (i.e., the original graph $(\mathbf{X}, \mathbf{A})$ is the input of the encoder $f_{\theta}$). 

After that, we obtain node-feature reconstruction errors $\mathbb{L}_{\mathbf{X}}(\mathcal{G}~\vert~f_{\theta}, g_{\psi}, h_{\phi})$ on each node
and adjacency-matrix reconstruction errors $\mathbb{L}_{\mathbf{A}}(\mathcal{G}~\vert~f_{\theta}, g_{\psi}, h_{\phi})$ on each node pair.
Formally,
\begin{align}
    \mathbb{L}_{\mathbf{X}}(\mathcal{G}) = 
    \mathbb{L}_{\mathbf{X}}(\mathcal{G}~\vert~f_{\theta}, g_{\psi}, h_{\phi}) &\coloneqq 
    \left[1 - \frac{\mathbf{X}_{i,:}^{T}\hat{\mathbf{X}}_{i,:}}{\lVert \mathbf{X}_{i,:} \rVert_{2} \cdot \lVert \hat{\mathbf{X}_{i,:}} \rVert}_{2}\right]_{i \in [\vert \mathcal{V}\vert]} \in \mathbb{R}^{\vert \mathcal{V}\vert},~\text{and}\label{eq:featureset}\\
    \mathbb{L}_{\mathbf{A}}(\mathcal{G}) = 
    \mathbb{L}_{\mathbf{A}}(\mathcal{G}~\vert~f_{\theta}, g_{\psi}, h_{\phi}) &\coloneqq 
    [\mathbf{A}_{i,j}\log{\hat{\mathbf{A}}_{i,j}} + (1 - \mathbf{A}_{i,j})\log{(1 - \hat{\mathbf{A}}_{i,j})}]_{i,j \in [\vert \mathcal{V}\vert]} \in \mathbb{R}^{\vert \mathcal{V}\vert^2}. \label{eq:edgeset}
\end{align}
Lastly, based on the motivation described in Section~\ref{sec:implication}, we aggregate reconstruction errors (i.e., ``\textit{summarize}'' vectors into scalars) using $T$ different aggregation functions $\texttt{Agg}_{t}$'s as follows:
\begin{equation}\label{eq:mainerrorvector}
    \texttt{Err}(\mathcal{G}) \coloneqq [\underbrace{\texttt{Agg}_{1}(\mathbb{L}_{\mathbf{X}}(\mathcal{G})), \cdots , \texttt{Agg}_{T}(\mathbb{L}_{\mathbf{X}}(\mathcal{G}))}_{\text{different aggregations of $\mathbb{L}_{\mathbf{X}}(\mathcal{G})$}},~ 
    \underbrace{\texttt{Agg}_{1}(\mathbb{L}_{\mathbf{A}}(\mathcal{G})) , \cdots , \texttt{Agg}_{T}(\mathbb{L}_{\mathbf{A}}(\mathcal{G}))}_{\text{different aggregations of $\mathbb{L}_{\mathbf{A}}(\mathcal{G})$}}] \in \mathbb{R}^{2T}.
\end{equation}
Any aggregation functions that provides a representative summary of $\mathbb{L}_{\mathbf{X}}(\mathcal{G})$ and $\mathbb{L}_{\mathbf{A}}(\mathcal{G})$ are applicable, such as mean and standard deviation.
As a result, we represent a graph $\mathcal{G}$ as a $2T$-dimensional vector $\texttt{Err}(\mathcal{G})$ (Eq.~(\ref{eq:mainerrorvector})), which we refer to as the final \textit{error representation} of $\mathcal{G}$.
We present a time complexity analysis of \method in Appendix~\ref{subsec:timecomplexity}.

After obtaining error representations of graphs, we leverage one-class classifiers on these representations to finally perform anomaly detection.
Note that Steps 1 and 2 are decoupled from the application of one-class classifiers (i.e., one-class classifier is trained after completing Steps 1 and 2).

\vspace{-1mm}
\section{Experiments}
\vspace{-1mm}
\label{sec:experiment}
\begin{table}[t]
\vspace{-6mm}
\caption{\textbf{GLAD performance}: Mean and standard deviation of test AUROC values ($\times$100) in the GLAD task are reported.
The \textcolor{sunwoogreen2}{best} and \textcolor{sunwooyellow2}{second-best} performances are highlighted in \textcolor{sunwoogreen2}{green} and \textcolor{sunwooyellow2}{yellow}.
A.R. denotes average ranking.
\method obtains the best average ranking among 18 methods.}
\setlength{\tabcolsep}{2.5pt}
\small
\centering
\scalebox{0.82}{
\renewcommand{\arraystretch}{1.0}
\begin{tabular}{c | l|c c c c c c c c c c | c}
    \toprule
        & Method & DD & Protein & NCI1 & AIDS & Reddit & IMDB & MUTAG & DHFR & BZR & ER &  AR \\
    \midrule 
    \midrule 
    \multirow{7}{*}{\rotatebox[origin=c]{90}{\textbf{GLAD methods}}}
     
     & DOMINANT-G~\citep{ding2019deep} & 64.3 {\std (4.4)} & 55.9 {\std (9.7)} & 65.5 {\std (6.1)} & 80.6 {\std (4.0)} & 58.6 {\std (5.3)} & 60.8 {\std (6.7)} & 65.0 {\std (4.2)} & 56.6 {\std (9.2)} & \best 76.2 {\std (7.8)} & 58.7 {\std (5.5)} & 10.7 \\
     
     & OCGTL~\citep{qiu2022raising} & 74.5 {\std (5.1)} &  71.0 {\std (8.7)} & 61.2 \std{(5.5)} & 95.3 {\std (3.7)} & 69.0 {\std (4.0)} & 65.8 {\std (5.8)} & 64.9 {\std (4.9)} & \best 66.5 {\std (9.9)} & 71.3 {\std (17.1)} & 63.0 {\std (3.6)} &  6.9 \\
     
     & GLocalKD~\citep{ma2022deep} & 47.8 {\std (8.5)} & 50.7 {\std (8.5)} & 51.6 {\std (5.6)} & 51.2 {\std (1.2)} & 49.8 {\std (4.2)} & 58.5 {\std (6.7)} & 55.1 {\std (4.4)} & 54.1 {\std (8.1)} &  55.8 {\std (16.7)} & 54.4 {\std (4.4)} &  17.0 \\
     
     & GLADC~\citep{luo2022deep} & 52.1 {\std (5.2)} & 50.7 {\std (5.6)}  & 51.4 {\std (3.6)} &  51.4 {\std (1.0)} & 52.2 {\std (2.6)} & 57.7 {\std (5.2)} & 53.3 {\std (4.5)} & 55.8 {\std (4.1)} & 59.0 {\std (14.5)} & 52.8 {\std (4.2)}  &  16.8 \\
     
     & GLAM~\citep{zhao2022graph} & 61.6 {\std (5.2)} & 60.3 {\std (5.6)} & 58.1 {\std (1.9)} & 93.6 {\std (2.6)} & 75.6 {\std (4.0)} & 65.1 {\std (3.5)} & 63.0 {\std (2.0)} & 57.2 {\std (2.7)} & 72.6 {\std (8.9)} & 55.2 {\std (2.9)} &  9.8 \\
     
     & HIMNET~\citep{niu2023graph} & 52.1 {\std (3.7)} & 56.9 {\std (5.8)} &  53.6 {\std (4.6)} &  64.3 {\std (3.2)}  & 65.7 {\std (2.4)} & 61.8 {\std (4.3)} & 57.5 {\std (2.9)} & 63.6 {\std (6.7)} & 72.0 {\std (9.9)} & 55.7 {\std (2.8)} &  12.3 \\
     
     & SIGNET~\citep{liu2024towards} & 64.2 {\std (9.3)} & 56.4 {\std (6.4)} & 63.1 {\std (4.0)} &  97.2 {\std (1.6)} & \secb 78.0 {\std (4.4)} & 48.2 {\std (4.8)} & 67.5 {\std (1.6)} & 40.2 {\std (5.8)} & 66.6 {\std (9.5)} & 56.2 {\std (4.3)} & 10.4 \\
     
     \midrule 

     \multirow{6}{*}{\rotatebox[origin=c]{90}{\textbf{SSL- based}}} 
     
     & GraphCL-1~\citep{you2020graph} & 64.5 {\std (3.9)} & 60.7 {\std (4.2)} & 55.8 {\std (3.1)} & 71.2 {\std (6.6)} & 57.7 {\std (5.5)} & 54.2 {\std (6.2)} & 53.6 {\std (2.3)} & 57.8 {\std (6.7)} & 60.5 {\std (9.3)} &  55.5 {\std (4.1)} &   14.2 \\
     
     & GAE-1~\citep{kipf2016variational} & 64.7 {\std (5.2)} & 61.3 {\std (7.0)} & 62.5 {\std (2.2)} & 86.2 {\std (1.4)} & 74.8 {\std (3.2)} & 63.8 {\std (7.4)} & 63.2 {\std (3.3)} & 56.5 {\std (9.6)} & 68.5 {\std (13.7)} & 60.0 {\std (3.9)} &  10.3 \\
     
     & GraphMAE-1~\citep{hou2022graphmae} & 56.7 {\std (7.3)} & 60.5 {\std (4.9)} & 53.4 {\std (3.2)} & 91.8 {\std (5.3)} & 72.7 {\std (3.2)} & 67.0 {\std (5.0)} & 62.3 {\std (2.6)} & 62.2 {\std (9.6)} & 70.1 {\std (7.6)} & 52.2 {\std (3.6)}  &  10.6 \\
     
     \cmidrule{2-13} 
     
    & GraphCL-2~\citep{you2020graph} & 66.1 {\std (3.0)} & 59.1 {\std (5.2)} & 60.3 {\std (4.4)} & 91.8 {\std (3.5)} & 77.3 {\std (4.1)} & 66.3 {\std (5.6)} & 67.4 {\std (3.3)} & 59.1 {\std (4.6)} &  71.9 {\std 10.4} & 67.3 {\std (3.4)} &  7.2 \\
     
     & GAE-2~\citep{kipf2016variational} & 67.2 {\std (3.4)} & 62.3 {\std (5.0)} & 62.4 {\std (3.9)} & 85.8 {\std (1.6)} & 75.3 {\std (5.7)} & 66.6 {\std (7.6)} & 67.3 {\std (3.3)} & 60.8 {\std (5.6)} & 72.0 {\std (8.8)} & 65.7 {\std (2.0)} &  7.0 \\
     
     & GraphMAE-2~\citep{hou2022graphmae} & 68.0 {\std (4.3)} & 61.2 {\std (4.0)} &  68.3 {\std (3.6)} & 90.8 {\std (3.6)} & 75.8 {\std (4.8)} & 66.7 {\std (5.8)} & \best 68.1 {\std (2.4)} & 61.4 {\std (6.0)} & \secb 72.8 {\std (6.4)} & 66.2 {\std (6.4)} &  \secb 5.1 \\
     
     \midrule 

     \multirow{4}{*}{\rotatebox[origin=c]{90}{\textbf{Variants}}} 

     & \method w/o $\mathbb{L}_{\mathbf{X}}$ & \secb 79.4 {\std (3.7)} & \secb 75.6 {\std (3.7)} & \secb 69.2 {\std (3.7)} & \secb 99.6 {\std (0.5)} & 72.2 {\std (4.0)} & 65.8 {\std (5.7)} & 65.8 {\std (3.1)} & 60.4 {\std (6.6)} & 65.6 {\std (19.4)} & 66.3 {\std (3.6)} & 5.8\\
     & \method w/o $\mathbb{L}_{\mathbf{A}}$ & 61.8 {\std (7.6)} & 64.7 {\std (7.1)} & 63.1 {\std (3.3)} & 89.3 {\std (2.8)} & 72.0 {\std (4.8)} & 56.9 {\std (7.1)} & 57.0 {\std (3.5)} & 58.1 {\std (3.1)} & 68.7 {\std (14.2)} & 60.7 {\std (4.0)} & 11.0\\
     & \method w/o \texttt{AVG} & 78.6 {\std (4.0)} & 68.1 {\std (5.5)} & 68.0 {\std (2.0)} & 95.0 {\std (2.6)} & 73.2 {\std (6.6)} & 66.2 {\std (6.5)} & 60.9 {\std (3.9)} & 60.1 {\std (2.4)} & 66.3 {\std (13.0)} & 62.0 {\std (3.5)} & 7.7\\
     & \method w/o \texttt{STD} & 74.3 {\std (5.4)} & 74.4 {\std (5.2)} & 65.2 {\std (3.6)} & 98.7 {\std (0.5)} & 70.5 {\std (4.3)} & \best 70.7 {\std (3.7)} & 62.0 {\std (2.4)} & 62.9 {\std (6.4)} & 71.3 {\std (11.5)} & \secb 66.7 {\std (2.4)} & 5.6\\

     \midrule 
     
     & \method & \best 80.5 {\std (2.3)} & \best 78.4 {\std (2.2)} & \best 71.1 {\std (2.0)} & \best 99.7 {\std (0.5)} & \best 78.4 {\std (5.7)} & \secb 69.2 {\std (3.5)} & \secb 67.5 {\std (3.4)} & \secb 63.8 {\std (8.6)} & 69.5 {\std (12.6)}  & \best 67.9 {\std (3.6)} &  \best 2.2 \\
     
     \bottomrule
\end{tabular}
\label{tab:mainexp}
}
\vspace{2mm}
\end{table}

In this section, we evaluate the effectiveness of \method in graph-level anomaly detection (GLAD) by addressing four key research questions.
\begin{itemize}[leftmargin=*]
    \item \textbf{RQ1.} How accurately does \method detect anomalous graphs? 
    \item \textbf{RQ2.} How robust is \method against contamination of the training set? 
    \item \textbf{RQ3.} Can the error representations of \method distinguish anomalies from normal graphs in the representation space? 
    \item \textbf{RQ4.} Are the key components of \method essential for its performance?
\end{itemize}

\subsection{Experimental settings}\label{subsec:expsetting}

\textbf{Datasets.}
Given the absence of benchmark datasets with ground-truth graph-level anomalies, following existing GLAD studies~\citep{qiu2022raising, liu2024towards, luo2022deep, zhang2022unsupervised, ma2022deep, niu2023graph}, we use graph classification benchmark datasets for evaluation.
Specifically, we use 10 datasets from diverse domains, such as chemical molecules, bioinformatics, and social networks. 
Detailed descriptions of the datasets are in Appendix~\ref{sec:datasets}.

\textbf{Training and evaluation.}
Following \citet{qiu2022raising}, for a dataset with $C$ graph classes, we use $C$ experimental configurations. 
{In each configuration, graphs with one class are treated as normal, while graphs with the other classes are considered anomalies.}
For each configuration, the normal graphs are split into training, validation, and test sets in an 80\%/10\%/10\% ratio.
Additionally, 5\% of anomalies are sampled for the validation set (only for hyperparameter tuning) and 5\% for the test set.
We conduct five trials for each configuration, {with each trial using different data splits and model initializations}.
Each model's performance on each dataset is evaluated by averaging the test mean AUROC across all configurations.

\textbf{Baseline methods and \method.}
We compare \method against 13 baseline methods, including 7 GLAD methods~\citep{qiu2022raising, liu2024towards, luo2022deep, zhang2022unsupervised, ma2022deep, niu2023graph, ding2019deep} and 6 graph self-supervised learning (SSL) methods~\citep{hou2022graphmae, kipf2016variational, you2020graph}.
Among SSL methods, those with `-1' in their names are one-stage SSL methods that use SSL losses as anomaly scores.
Those with `-2' are two-stage SSL methods that (1) obtain graph embeddings the models learn via SSL and (2) employ a one-class classifier.
For aggregation functions of \method (Eq.~(\ref{eq:mainerrorvector})), we use two aggregation functions: mean and standard deviation. 
Therefore, we represent each graph with a $4$-dimensional vector, which is formalized in Appendix~\ref{subsec:aggregationdetail}.
All methods use GIN~\citep{xu2018powerful} as a graph encoder. 
All two-stage approaches use an MLP autoencoder as a one-class classifier. 
Details of two-stage methods and the MLP autoencoder are in Appendix~\ref{subsec:twostagedetails}.
In addition, we provide further experimental details, including hyperparameter configurations, in Appendix~\ref{sec:experimentdetails}.

\vspace{-2.5mm}
\subsection{RQ1. Performance in graph-level anomaly detection}
As shown in Table~\ref{tab:mainexp}, \method outperforms all baseline methods in terms of average ranking. 
Two points stand out.
First, in 8 out of 10 datasets, \method outperforms all other GLAD methods. 
Compared to the second-best GLAD method (OCGTL), \method achieves up to 16.2\% performance gain (in the NCI1 dataset).
Second, in 8 out of 10 datasets, \method outperforms all the SSL-based two-stage baselines (GraphCL-2, GAE-2, and GraphMAE-2). 
Given that they use the same one-class classifier with \method, this result demonstrates the effectiveness of the proposed error representation (Eq.~(\ref{eq:mainerrorvector})).

\subsection{RQ2. Robustness against the training set contamination}

\begin{figure}[t] 
    \centering
    \includegraphics[width=1.0\linewidth]{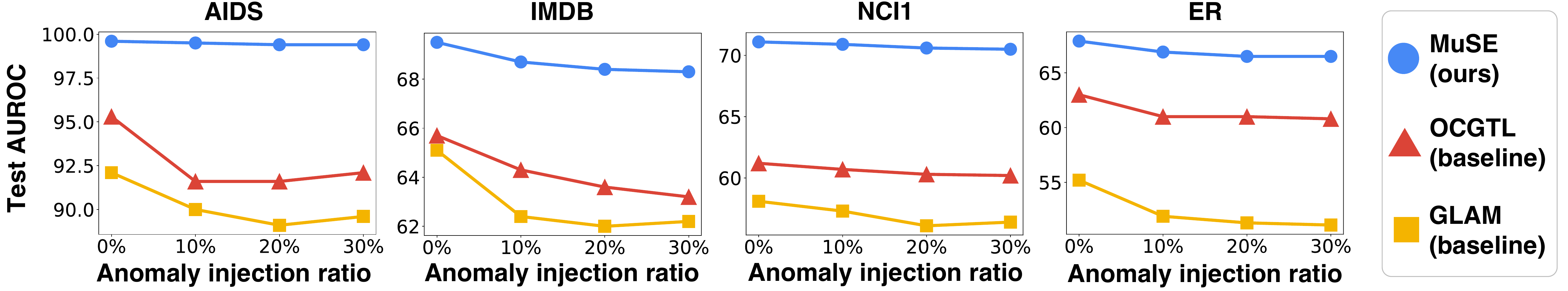}

    \caption{\label{fig:resultrobust}
    Comparison of the three strongest GLAD methods' robustness against training set contamination. \method undergoes the least performance drop among the three methods.
    }
\end{figure}

In real-world scenarios, training data may also contain anomalous graphs, making the robustness of GLAD methods against training set contamination crucial for their practical effectiveness.
To assess this robustness, we inject anomalous graphs into the training set at rates of 10\%, 20\%, and 30\%.~\footnote{Specifically, for a training set containing $n$ graphs, we sample $n \times (k/100)$ anomalous graphs and inject.}
As shown in Figure~\ref{fig:resultrobust}, \method exhibits the least performance drop, among the three strongest GLAD methods, demonstrating its robustness. 

\subsection{RQ3. Error representation visualization}
\begin{wrapfigure}{r}{0.6\linewidth}
    \centering
    \vspace{-8mm}
    \includegraphics[width=\linewidth]{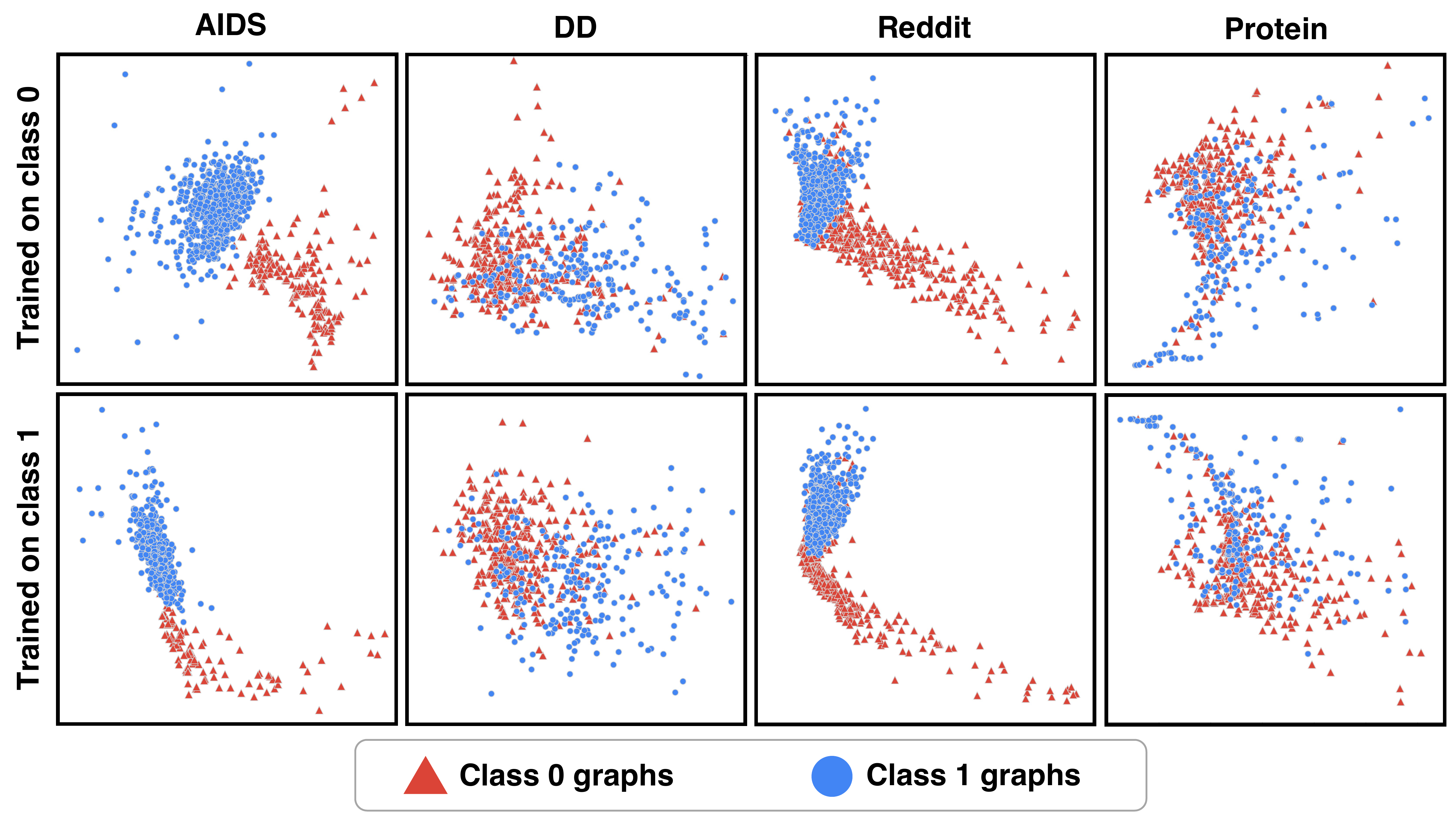}
    \caption{\label{fig:resultembeddings}PCA visualization of \method's error representations of graphs.
Graphs belonging to different classes are well-separated in the representation space.}
    \vspace{-10mm}
\end{wrapfigure}
We evaluate whether the error representation offered by \method can effectively distinguish graphs of different classes.
We sample half of the graphs belonging to the same class to train \method.
Then, we visualize the error representations generated by the trained \method for the rest half and graphs sampled from another class. 
As shown in Figure~\ref{fig:resultembeddings}, the error representations produced by \method are distinguishable across different classes.

\subsection{RQ4. Ablation study}
Lastly, we provide an ablation study of \method.
We aim to justify the two key components of \method: (1) the reconstruction errors $\mathbb{L}_{\mathbf{X}}$ and $\mathbb{L}_{\mathbf{A}}$, and (2) multiple aggregation functions (specifically, mean and standard deviation).
To this end, we leverage the following four variants of \method: 
\begin{enumerate}[leftmargin=*, itemsep=0.0cm, start = 1, label={(\bfseries V\arabic*)}]
    \item \method w/o $\mathbb{L}_{\mathbf{X}}$: A variant that only leverages adjacency matrix reconstruction,
    \item \method w/o $\mathbb{L}_{\mathbf{A}}$: A variant that only leverages node feature reconstruction,
    \item \method w/o \texttt{STD}: A variant that only leverages mean aggregation,
    \item \method w/o \texttt{AVG}: A variant that only leverages standard-deviation aggregation.
\end{enumerate}
As shown in Table~\ref{tab:mainexp}, \method with all components outperforms all of its variants in 8 out of 10 datasets, justifying the design choice of \method.

    


\section{Discussions}
\label{sec:conclusion}
In this work, we report and analyze an intriguing phenomenon, \textit{reconstruction flip} of graph autoencoders (Graph-AEs).
We investigate the phenomenon theoretically and empirically (Section~\ref{sec:analysis}) and further claim their implication in graph-level anomaly detection (GLAD; Section~\ref{sec:implication}).
Based on our analysis, we propose a novel GLAD method, \method (Section~\ref{sec:method}), and report its superior performance in GLAD via extensive experiments (Section~\ref{sec:experiment}).
Below, we conclude the paper by discussing some potential limitations of our research, which could be addressed in future work.

{\textbf{More diverse and general primary patterns.}
In our analysis, we focus on the two patterns: 
\textit{community structures} and \textit{a node cycle} (Section~\ref{sec:analysis}).
Although our claims hold for these patterns, a wide range of graph patterns remains unexplored in this work.
Furthermore, a unified framework that can incorporate various graph patterns into a single pattern may exist.
Further investigation into these patterns may improve our analysis, leading to new potential applications.}

{\textbf{Scalability improvement.}
Since \method reconstructs the entire adjacency matrix of a graph (Eq.~(\ref{eq:edgemainloss})), its time complexity for reconstruction is $O(n^{2})$, where $n$ is the number of nodes in the graph.
While this complexity is manageable for many real-world graph-level datasets (refer to Table~\ref{tab:datasets}), leveraging \method for large-scale graphs can be challenging.
A simple technique that reconstructs only a sampled subset of an adjacency matrix results in a performance drop (refer to Appendix~\ref{subsec:scalability}). 
Therefore, improving the scalability of~\method while preserving its detection performance is a non-trivial task, making it a promising direction for future research.}

\newpage

\textbf{Acknowledgements.}
This work was supported by the National Research Foundation of Korea (NRF) grant funded
by the Korea government (MSIT) (No. RS-2024-00406985) (90\%).
This work was supported by Institute of Information \& Communications Technology Planning \& Evaluation (IITP) grant funded by the Korea government (MSIT)  (No. RS-2019-II190075, Artificial Intelligence Graduate School Program (KAIST)) (10\%).

\bibliographystyle{abbrvnat}
\bibliography{neurips_2024}

\newpage 

\appendix

\section{Proof}
\label{sec:appendix_proof}

In this section, we provide proof of each theorem presented in the paper.

\subsection{Proof of Theorem~\ref{theorem:generalization}}

\begin{reptheorem}{theorem:generalization}[Suspicious generalization across $\tau$]
        For any $n \geq 12$ and $0 < \tau_1 \leq 1$, 
    there exists $\epsilon > 0$ such that for any $0 < \gamma \leq \epsilon$ 
    and any $0 < \tau_2 \leq 1$
    with the initial weight matrix $\mathbf{W}^{(0)} = \mathbb{I}_{n}$,
    \begin{equation*}
        \underbrace{\mathbb{E}_{\mathcal{G}} [\mathcal{L}(\mathcal{G},\mathbb{W}(\tau_{1}, \mathbf{W}^{(0)}, \gamma)) \vert \tau_{2}]}_{\text{the error on graphs of strength $\tau_2$ after update}} < 
        \underbrace{\mathbb{E}_{\mathcal{G}} [\mathcal{L}(\mathcal{G},\mathbf{W}^{(0)}) \vert \tau_{2}]}_{\text{the error on graphs of strength $\tau_2$ before update}}.
    \end{equation*}
\end{reptheorem}
\begin{proof}
    Note that the reconstruction loss is defined as follows:
\begin{equation}\label{eq:proofequation}
    \mathcal{L} \coloneqq \lVert \mathbf{A} - \mathbf{AXW}(\mathbf{AXW})^{T}\rVert^{2}_{F}.
\end{equation}
Here, since $\mathbf{X}$ is identity, Eq.~(\ref{eq:proofequation}) is equal to $\mathcal{L} \coloneqq \lVert \mathbf{A} - \mathbf{A} \mathbf{W} \mathbf{W} \mathbf{A}\rVert^{2}_{F}$.
The derivative of Eq.~(\ref{eq:proofequation}) and its expectation are as follows~\citep{laue2018computing}:
\begin{align*}
    \left. \frac{\partial \mathcal{L}}{\partial \mathbf{W}}\right|_{\mathbf{W} = \mathbb{I}_{n}} &= 2(\mathbf{A}^{4} - \mathbf{A}^{3}), \\
    \mathbb{E}_{\mathbf{A}}\left[ \left.\frac{\partial \mathcal{L}}{\partial \mathbf{W}}\right|_{\mathbf{W} = \mathbb{I}_{n}} \right] &= 2( \mathbb{E}[\mathbf{A}^{4}] - \mathbb{E}[\mathbf{A}^{3}]).
\end{align*}

    Given $n$ (the total number of nodes) and $\tau$ (the community strength), let $N = \frac{n}{2}$ (the number of nodes in each community) and $p = \frac{1 + \tau}{2}$ (the inner-community edge probability), we have
    
    \textbf{(1) If $i = j$:}
    \begin{align*}
    \mathbb{E}[\mathbf{A}^3_{ij}] 
    &= 
    (N - 1) (N - 2) p^3 +  
    2 N (N - 1) p (1 - p)^2 +  
    N (N - 1) p (1 - p)^2,  
    \\
    &= 
    (N - 1) (N - 2) p^3 + 
    3 N (N - 1) p (1 - p)^2 ,   
    \end{align*}
    and
    \begin{align*}
    \mathbb{E}[\mathbf{A}^4_{ij}] 
    &= (N - 1) (N - 2) (N - 3) p^4 + 
    (N - 1) (N - 2) p^2 + 
    (N - 1) (N - 2) p^2 + 
    (N - 1) p 
    \\&~~~~+
    2 (N - 1) (N - 2) N p^2 (1 - p)^2 + 
    2 N (N - 1) p (1 - p)^2 
    \\&~~~~+
    (N - 1) (N - 2) N p^2 (1 - p)^2 + 
    (N - 1) N p (1 - p) 
    \\&~~~~+
    2 N^2 (N - 1) p^2 (1 - p)^2 + 
    N (N - 1)^2 (1 - p)^4 + 
    N (N - 1) (1 - p)^2 + 
    N (N - 1) (1 - p)^2 
    N (1 - p) 
    \\&~~~~+
    N (N - 1) (N - 2) p^2 (1 - p)^2 + 
    N (N - 1) p (1 - p), 
    \\
    &=
    (-6 + 11 N - 6 N^2 + N^3) p^4 + 
    2 (-1 + N) p^2 (-2 + N - 4 N (1-p)^2 + 3 N^2 (1-p)^2)
    \\&~~~~+
    N (1-p) (1 + 2 (-1 + N) (1-p) + (-1 + N)^2 (1-p)^3) 
    \\&~~~~+
    p (-1 + 2 N^2 (1-p) (1 + (1-p)) + N (1 - 2 (1-p) - 2 (1-p)^2)),
\end{align*}

    \textbf{(2) If $i$ and $j$ are in the same group:}
    \begin{align*}
    \mathbb{E}[\mathbf{A}^3_{ij}] 
    &= 
    (N - 2) (N - 3) p^3 +  
    p +  
    2 N (N - 2) p (1 - p)^2 +  
    2 N p (1 - p) +  
    N (N - 1) p (1 - p)^2  , 
    \\
    &= (N - 2) (N - 3) p^3 + 
    N (3N - 5) p (1 - p)^2 +
    p + 2 N p (1 - p),
    \end{align*}
    and
    \begin{align*}
    \mathbb{E}[\mathbf{A}^4_{ij}] 
    &=
    (N - 2) (N - 3) (N - 4) p^4 + 
    (N - 2) (N - 3) p^3 + 
    2 (N - 2) (N - 3) p^3 + 
    2 (N - 2) p^3 
    \\&~~~~+
    (N - 2) p^3 + 
    2 (N - 2) (N - 3) p^4 + 
    2 (N - 2) p^2 
    \\&~~~~+
    2 (N - 2) (N - 3) N p^2 (1 - p)^2 + 
    2 (N - 2) N p (1 - p)^2 + 
    2 N p (1 - p)^2 + 
    2 (N - 2) N p^2 (1 - p)^2 
    \\&~~~~+
    (N - 2) (N - 3) N p^2 (1 - p)^2 + 
    2 (N - 2) N p^2 (1 - p)^2 + 
    N p (1 - p)^2 + 
    2 N (N - 1) (N - 2) p^2 (1 - p)^2 
    \\&~~~~+
    2 N (N - 1) p^2 (1 - p)^2 + 
    N (N - 1) (N - 2) (1 - p)^4 + 
    N (N - 2) (1 - p)^3 + 
    2 N (N - 1) (1 - p)^4 
    \\&~~~~+
    2 N (1 - p)^2 + 
    N (N - 1) (N - 2) p^2 (1 - p)^2 + 
    N (N - 1) p (1 - p)^2 , 
    \\
    &= 
    3 (-2 + N)^2 p^3 + 
    (-3 + N) (-2 + N)^2 p^4 + 
    N (-2 + 3 N) p (1-p)^2 
    \\&~~~~+
    N (1-p)^2 (2 + (-2 + N) (1-p) + (-1 + N) N (1-p)^2) 
    \\&~~~~+
    2 p^2 (-2 + N + 7 N (1-p)^2 - 9 N^2 (1-p)^2 + 3 N^3 (1-p)^2).
\end{align*}

    \textbf{(3) If $i$ and $j$ are in different groups}
    \begin{align*}
    \mathbb{E}[\mathbf{A}^3_{ij}] 
    &= 
    2 (N - 1) (N - 2) p^2 (1 - p) +  
    2 (N - 1) p (1 - p)  
    \\&~~~~+
    (N - 1)^2 p^2 (1 - p) +  
    (N - 1)^2 (1 - p)^3 + 
    (1 - p),  
    \\
    &= 
    (N - 1) (3N - 5) p^2 (1 - p) +
    (N - 1)^2 (1 - p)^3 +
    2 (N - 1) p (1 - p) + (1 - p),
    \end{align*}
    and
    \begin{align*}
    \mathbb{E}[\mathbf{A}^4_{ij}] 
    &=
    (N - 1) (N - 2) (N - 3) p^3 (1 - p) + 
    (N - 1) (N - 2) p^3 (1 - p) + 
    (N - 1) (N - 2) p^2 (1 - p) 
    \\&~~~~+
    (N - 1) (N - 2) p^2 (1 - p) + 
    (N - 1) p (1 - p) + 
    (N - 1)^2 (N - 2) p^3 (1 - p) 
    \\&~~~~+
    (N - 1) (N - 2) p^2 (1 - p) + 
    (N - 1)^2 (N - 2) p (1 - p)^3 + 
    (N - 1)^2 p (1 - p)^3 
    \\&~~~~+
    (N - 1) (N - 2) p (1 - p)^2 + 
    (N - 1) p (1 - p)^2 + 
    (N - 1)^2 p (1 - p)^2 + 
    (N - 1) p (1 - p) 
    \\&~~~~+
    (N - 1)^2 (N - 2) p^3 (1 - p) + 
    (N - 1)^2 p^2 (1 - p) + 
    (N - 1)^2 (N - 2) p (1 - p)^3 
    \\&~~~~+
    (N - 1)^2 p (1 - p)^2 + 
    (N - 1) (N - 2) p (1 - p)^3 + 
    (N - 1)^2 p (1 - p)^3 + 
    (N - 1) p (1 - p)^2 
    \\&~~~~+
    (N - 1) p (1 - p)^2 + 
    (N - 1)^2 (N - 2) p (1 - p)^3 + 
    (N - 1) (N - 2) p (1 - p)^2 
    \\&~~~~+
    (N - 1)^2 p (1 - p)^3 + 
    (N - 1)^2 p (1 - p)^2 + 
    (N - 1) p (1 - p)^2 + 
    (N - 1) p (1 - p) 
    \\&~~~~+
    (N - 1)^2 (N - 2) p (1 - p)^3 + 
    (N - 1)^2 p (1 - p)^3 + 
    (N - 1)^2 p (1 - p)^3 + 
    (N - 1) (N - 2) p (1 - p)^3 
    \\&~~~~+
    (N - 1) p (1 - p)^2 + 
    (N - 1) p (1 - p)^2 + 
    (N - 1) (N - 2) (N - 3) p^3 (1 - p) 
    \\&~~~~+
    (N - 1) (N - 2) p^2 (1 - p) + 
    (N - 1) (N - 2) p^2 (1 - p) + 
    (N - 1) p (1 - p) 
    \\
    &= (-1 + N) p (1-p) \triangle,
\end{align*}
where
\[
\triangle = 4 + (-11 + 6 N) p + (14 - 15 N + 4 N^2) p^2 + (-1 + 
      5 N) (1-p) + (-1 - 5 N + 4 N^2) (1-p)^2.
\]

Hence, the entries in
\[
\mathbb{E}_{A}[\mathbf{A}^{4} - \mathbf{A}^{3}] = \frac{1}{2}\mathbb{E}_{\mathbf{A}}\left[ \left.\frac{\partial \mathcal{L}}{\partial \mathbf{W}}\right|_{\mathbf{W} = \mathbb{I}_{n}} \right]
\]
have three different values: 
(1) $i$ and $j$ are equal, 
(2) $i \neq j$ and $v_{i}$ and $v_{j}$ belong to the same group, and 
(3) $i \neq j$ and $v_{i}$ and $v_{j}$ belong to the different groups.
Thus, we can express $\mathbb{E}_{A}[\mathbf{A}^{4} - \mathbf{A}^{3}]$ as $a\mathbf{I} + b\mathbf{P} + c\mathbf{U}$, where $a,b,c \in \mathbb{R}$ and $\mathbf{P}$ and $\mathbf{U}$ are defined as:
\begin{equation}
    \mathbf{P} = \begin{pmatrix}
    \mathbf{1} & \mathbf{0} \\
    \mathbf{0} & \mathbf{1}
    \end{pmatrix},~\text{and}~
    \mathbf{U} = \begin{pmatrix}
    \mathbf{0} & \mathbf{1} \\
    \mathbf{1} & \mathbf{0}
    \end{pmatrix},
\end{equation}
where $\mathbf{1} \in \{1\}^{N \times N}$ and $\mathbf{0} \in \{0\}^{N \times N}$.

Specifically,
\begin{align*}
    a &= 3 (N - 2) (N - 3) p^4 + 
    2N (4N - 6) p (1 - p)^2 +
    N (N - 1) (1 - p)^4 \\    
    b &= (N - 2) (N - 3) p^3 ((N - 4) p - 1) + 
    N (3N - 5) p (1 - p)^2 (2 (N - 2) p - 1)
    \\&~~~~+
    N (N - 1) (N - 2) (1 - p)^4 ,\\
    &= 
    N^3 (1 - 4 p + 12 p^2 - 16 p^3 + 8 p^4) + 
    N^2 (-3 + 9 p - 34 p^2 + 52 p^3 - 34 p^4) 
    \\&~~~~+
    N (2 - 3 p + 22 p^2 - 38 p^3 + 48 p^4)
    -6 p^3 (1 + 4 p)
    \\
    c &= (N - 1) p^2 (1 - p) (4 (N - 2)^2 p - 3N + 5) + 
    (N - 1) (1 - p)^3 (4 (N - 2)^2 p - N + 1)
\end{align*}

For any training graphs with $0 < \tau_1 \leq 1$, i.e., $0.5 < p \leq 1$,
we have
\begin{itemize}[leftmargin=*]
    \item For any $N \geq 4$, $a > 0$,
    \item For any $N \geq 6$, $b > 0$,
    \item For any $N \geq 4$, $c > 0$.
\end{itemize}
Thus, $a, b, c > 0$ for any $N \geq 6$ (i.e., $n \geq 12$) hold.

Let $W' = \mathbb{W}(\tau_{1}, \mathbf{W}^{(0)}, \gamma)$ be the updated $W$ after one step of gradient descent with learning rate $\gamma$ trained on graphs with $\tau_1$ (and $n$).
Now, we derive the reconstruction loss on test graphs computed with $W'$.
Formally, reconstruction loss by using $W'$ is defined as follows:
\begin{equation}\label{eq:theoryloss}
    \mathcal{L} = \lVert \mathbf{A} - \mathbf{A}W'(\mathbf{A}W')^{T}\rVert^{2}_{2} 
    = \lVert \mathbf{A} - \mathbf{A}(W')^{2}\mathbf{A}\rVert^{2}_{2}.
\end{equation}
We further simplify $(W')^{2}$ as follows:
\begin{equation}\label{eq:updatedgradientsquare}
    (W')^{2} = \mathbf{I} + 2(\epsilon_{1}\mathbf{I} + \epsilon_{2}\mathbf{P} + \epsilon_{3}\mathbf{U}) + (\epsilon_{1}\mathbf{I} + \epsilon_{2}\mathbf{P} + \epsilon_{3}\mathbf{U})^{2},
\end{equation}
where
$\epsilon_1 = - \frac{1}{2} \gamma a$,
$\epsilon_2 = - \frac{1}{2} \gamma b$, and
$\epsilon_3 = - \frac{1}{2} \gamma c$.

To simplify Eq.~(\ref{eq:updatedgradientsquare}), we use the property that $\mathbf{P}^{2} = N\mathbf{P}$, $\mathbf{U}^{2} = N\mathbf{U}$, and $\mathbf{PU}=\mathbf{UP} = N\mathbf{U}$.
Using these facts, we rewrite Eq.~(\ref{eq:updatedgradientsquare}) as below:
\begin{equation}\label{eq:simplifiedweight}
    (W')^{2} = \mathbf{I} + 2(\epsilon_{1}\mathbf{I} + \epsilon_{2}\mathbf{P} + \epsilon_{3}\mathbf{U}) + (\epsilon^{2}_{1}\mathbf{I} + \epsilon^{2}_{2}N\mathbf{P} + \epsilon^{2}_{3}N\mathbf{U}) + 2\epsilon_{1}\epsilon_{2}\mathbf{P} + 2\epsilon_{1}\epsilon_{3}\mathbf{U} + 2\epsilon_{2}\epsilon_{3}N\mathbf{U},
\end{equation}
where
\begin{align*}
    \alpha_1 &= 2 \epsilon_1 + \epsilon^2_1 \\
    \alpha_2 &= 2 \epsilon_2 + N\epsilon^2_2 + 2\epsilon_1 \epsilon_2 \\
    \alpha_3 &= 2 \epsilon_3 + N\epsilon^2_3 + 2\epsilon_1 \epsilon_3 + 2N \epsilon_2 \epsilon_3
\end{align*}

Now, by using the derived results, we rewrite the reconstruction loss term Eq.~(\ref{eq:theoryloss}) as below:
\begin{equation}
    \mathcal{L} = \lVert \mathbf{A} - \mathbf{A}(\mathbf{I} + \alpha_{1} \mathbf{I} + \alpha_{2} \mathbf{P}  + \alpha_{3} \mathbf{P})\mathbf{A} \rVert^{2}_{F}
    = \lVert (\mathbf{A} - \mathbf{A}^{2}) - (\alpha_{1}\mathbf{A}^{2} + \alpha_{2}\mathbf{APA} + \alpha_{3}\mathbf{AUA})  \rVert^{2}_{F},
\end{equation}
where
\begin{align*}
    (\mathbf{APA})_{ij} &= \sum_{\text{$k, \ell$ in the same group}} \mathbf{A}_{ik} \mathbf{A}_{jl} \\
    (\mathbf{AUA})_{ij} &= \sum_{\text{$k, \ell$ in different groups}} \mathbf{A}_{ik} \mathbf{A}_{jl} \\
    (\mathbf{A}^2)_{ij} &= \sum_k \mathbf{A}_{ik} \mathbf{A}_{jk}
\end{align*}

We can decompose the loss by
\begin{equation*}
    \mathcal{L} = \sum_{i, j} \left(\mathbf{A}_{ij} - (1 + \alpha_1)(\mathbf{A}^2)_{ij}
    - \alpha_2 (\mathbf{APA})_{ij}
    - \alpha_3 (\mathbf{AUA})_{ij}\right)
    )^2.
\end{equation*}


Given $i$ and $j$, let $x = A_{ij}$, $y = (\mathbf{A}^2)_{ij}$, $z = (\mathbf{APA})_{ij}$, and $w = (\mathbf{AUA})_{ij}$,
we have
\begin{align*}
    \mathcal{L}_{ij} 
    &= (
    x - (1 + \alpha_1)y
    - \alpha_2 z
    - \alpha_3 w
    )^2 \\
    &= x^2 + 
    - 2 (1 + \alpha_1) x y     
    - 2 \alpha_2 x z 
    - 2 \alpha_3 x w
    + (1 + \alpha_1)^2 y^2
    \\&~~~~+
    2 (1 + \alpha_1) \alpha_2 y z     
    + 2 (1 + \alpha_1) \alpha_3 y w    
    + \alpha_2^2 z^2
    + 2 \alpha_2 \alpha_3 z w
    + \alpha_3^2 w^2.
\end{align*}

Let $\bar{\mathcal{L}}_{ij} = (\mathbf{A}_{ij} - (\mathbf{A}^2)_{ij})^2 = (x - y)^2.$
By using this notation, we derive the following result:
\begin{align*}
    \mathcal{L}_{ij} - \bar{\mathcal{L}}_{ij} 
    &= - 2 \alpha_1 x y
    - 2 \alpha_2 x z 
    - 2 \alpha_3 x w
    + \alpha_1 (\alpha_1 + 2) y^2
    \\&~~~~+
    2 (1 + \alpha_1) \alpha_2 y z     
    + 2 (1 + \alpha_1) \alpha_3 y w    
    + \alpha_2^2 z^2
    + 2 \alpha_2 \alpha_3 z w
    + \alpha_3^2 w^2.
\end{align*}
We shall show that 
\[
\mathbb{E} [\sum_{i, j} (\mathcal{L}_{ij} - \bar{\mathcal{L}}_{ij})] < 0.
\]

As $\gamma \to 0^+$, ignoring higher-order terms and keeping the term linear w.r.t. $\gamma$, it suffices to show that
\[
\sum_{i,j} a (\mathbb{E}[xy] - \mathbb{E}[y^2]) + b(\mathbb{E}[xz] - \mathbb{E}[yz]) + c(\mathbb{E}[xw] - \mathbb{E}[yw]) < 0.
\]
Note: In the following derivation, $p$ is from test graphs (i.e., $p = \frac{\tau_2 + 1}{2}$).

\textbf{(1) If $i = j$}
\begin{align*}
    \mathbb{E}[xy] &= 0\\
    \mathbb{E}[xz] &= 0\\
    \mathbb{E}[xw] &= 0\\
    \mathbb{E}[y^2] &= 
    (N - 1) p  
    + (N - 1) (N - 2) p^2 
    + N (1 - p) 
    + N (N - 1) (1 - p)^2 
    + 2 N (N - 1) p (1 - p) 
    \\
    &= N^2 +
    N (-2p^2) +
    (2p^2 - p)    
    \\
    \mathbb{E}[yz] &= 
    (N - 1) (N - 2) (N - 3) p^3 + 
    3 (N - 1) (N - 2) p^2 + 
    (N - 1) p 
    \\&~~~~+
    N (N - 1)^2 p (1-p)^2 + 
    N (N - 1) p (1-p) 
    \\&~~~~+
    N (N - 1) (N - 2) p^2 (1-p) + 
    N (N - 1) p (1 - p) 
    \\&~~~~+
    N (N - 1) (N - 2) (1-p)^3 + 
    3 N (N - 1) (1-p)^2 + 
    N (1 - p) 
    \\
    &= 
    N^3 (1 - 2 p + 2 p^2) + 
    N^2 (3p - 4 p^2 - 2 p^3)
    \\&~~~~+
    N (-p - 4 p^2 + 8 p^3) +
    (-p + 6 p^2 - 6 p^3)
    \\
    \mathbb{E}[yw] &= 
    2 N (N - 1) (N - 2) p^2 (1-p) + 
    2 N (N - 1) p (1-p) 
    \\&~~~~+
    2 N (N - 1)^2 p (1-p)^2 + 
    2 N (N - 1) p (1-p) 
    \\
    &=
    N^3 (2p - 2p^2) + 
    N^2 (2p^3 - 2p^2) +
    N (4 p^2 -2p - 2p^3)
\end{align*}

Hence,
\begin{align*}
    d_{11} := \mathbb{E}[xy] - \mathbb{E}[y^2] &= -N^2 + N(2p^2) + (p-2p^2) \\
    d_{12} := \mathbb{E}[xz] - \mathbb{E}[yz] &= N^3 (-1 + 2 p - 2 p^2) + 
    N^2 (-3p + 4 p^2 + 2 p^3)
    \\&~~~~+
    N (p + 4 p^2 - 8 p^3) +
    (p - 6 p^2 + 6 p^3) \\
    d_{13} :=\mathbb{E}[xw] - \mathbb{E}[yw] &= 
    N^3 (-2p + 2p^2) + 
    N^2 (-2p^3 + 2p^2) +
    N (-4 p^2 +2p + 2p^3)
\end{align*}

\textbf{(2) Else if $i$ and $j$ are in the same group}

\begin{align*}
    \mathbb{E}[xy] &= p \mathbb{E}[y] = N p (1 - 2p + 2p^2) - 2 p^3 \\
    \mathbb{E}[xz] &= p \mathbb{E}[z] = N^2 p (1 - 2p + 2p^2) - 2 N p^3 + p^3 \\
    \mathbb{E}[xw] &= p \mathbb{E}[w] = 2 N^2 (p^2 - p^3) + 4 N (p^3 - p^2) \\
    \mathbb{E}[y^2] &=  
    (N - 2) p^2 
    + (N - 2) (N - 3) p^4 
    + N (1 - p)^2 
    + N (N - 1) (1 - p)^4 
    + 2 N (N - 2) p^2 (1 - p)^2 
    \\
    &=
    N^2 (1 - 2 p + 2 p^2)^2 +
    2 N p (1 - 4 p + 6 p^2 - 5 p^3) +
    6 p^4 -2 p^2
    \\
    \mathbb{E}[yz] &=     
    (N - 2) (N - 3) (N - 4) p^4 + 
    2 (N - 2) (N - 3) p^4 +       
    (N - 2) p^3                   
    \\&~~~~+
    (N - 2) (N - 3) p^4 +         
    2 (N - 2) (N - 3) p^3 +       
    (N - 2) p^2                   
    \\&~~~~+
    N (N - 1) (N - 2) p^2 (1-p)^2 +     
    N (N - 2) p^2 (1-p)                 
    \\&~~~~+
    N (N - 2) (N - 3) p^2 (1-p)^2 +   
    N (N - 2) p (1-p)^2               
    \\&~~~~+
    N (N - 1) (N - 2) (1-p)^4 +     
    N (N - 1) (1-p)^4 +             
    2 N (N - 1) (1-p)^3 +           
    N (1-p)^2                       
    \\
    &= 
    N^3 (1 - 2 p + 2 p^2)^2 - 
    2 p^2 (1 - 5 p + 3 p^2) 
    \\&~~~~+
    N^2 p (3 - 15 p + 24 p^2 - 16 p^3) + 
    N p (-2 + 12 p - 27 p^2 + 20 p^3)
    \\
    \mathbb{E}[yw] &= 
    2 N (N - 2) (N - 3) p^3 (1-p) + 
    2 N (N - 2) p^3 (1-p) + 
    2 N (N - 2) p^2 (1-p)   
    \\&~~~~+
    2 N (N - 1) (N - 2) p (1-p)^3 + 
    2 N (N - 2) p (1-p)^3 + 
    2 N (N - 2) p (1-p)^2 
    \\
    &=
    N^3 (2 p - 6 p^2 + 8 p^3 - 4 p^4) +
    N^2 (-2 p + 10 p^2 - 20 p^3 + 12 p^4) +
    N (-4 p + 4 p^2 + 8 p^3 - 8 p^4)
\end{align*}

Hence,
\begin{align*}
    d_{21} := \mathbb{E}[xy] - \mathbb{E}[y^2] &= 
    -N^2 (1 - 2 p + 2 p^2)^2 +
    N p (-1 + 6 p - 10 p^2 + 10 p^3) +
    2 p^2 (1 - p - 3 p^2)  \\
    d_{22} := \mathbb{E}[xz] - \mathbb{E}[yz] &= 
    -N^3 (1 - 2 p + 2 p^2)^2 + p^2 (2 - 9 p + 6 p^2) 
    \\&~~~~+
    N p (2 - 12 p + 25 p^2 - 20 p^3) + 
    N^2 p (-2 + 13 p - 22 p^2 + 16 p^3)
    \\
    d_{23} := \mathbb{E}[xw] - \mathbb{E}[yw] &= 2 N (-1 + p) p (N (-1 + 3 p - 6 p^2) + N^2 (1 - 2 p + 2 p^2) + 
   2 (-1 + p + 2 p^2))
\end{align*}

\textbf{(3) Else if $i$ and $j$ are in different groups}

\begin{align*}
    \mathbb{E}[xy] &= (1 - p) \mathbb{E}[y] = 2(N-1)p(1-p)^2\\
    \mathbb{E}[xz] &= (1 - p) \mathbb{E}[z] = 2 (N - 1)^2 p (1 - p)^2 \\
    \mathbb{E}[xw] &= (1 - p) \mathbb{E}[w] = (N - 1)^2 p^2 (1-p) + N^2 (1 - p)^3 \\
    \mathbb{E}[y^2] &= 
    2(N - 1) p (1 - p) + 
    2(N - 1) (N - 2) p^2 (1 - p)^2 + 
    2(N - 1)^2 p^2 (1 - p)^2 
    \\
    &= 2(N - 1) p (1 - p) + 2(N - 1) (2N - 3) p^2 (1 - p)^2
    \\
    \mathbb{E}[yz] &= 
    2 (N - 1) (N - 2) (N - 3) p^2 (1-p)^2 +     
    2 (N - 1) (N - 2) p^2 (1-p)^2               
    \\&~~~~+
    2 (N - 1) (N - 2) p^2 (1-p)^2 +             
    2 (N - 1) (N - 2) p (1-p)^2                 
    \\&~~~~+
    2 (N - 1) (N - 2) p^2 (1-p) +               
    2 (N - 1) p (1-p)                           
    \\&~~~~+
    2 (N - 1)^2 (N - 2) p^2 (1-p)^2 +     
    2 (N - 1)^2 p^2 (1-p)^2               
    \\
    &=
    N^3 (4p^2 - 8p^3 + 4p^4) +
    N^2 (2p - 16p^2 + 28p^3 - 14p^4)
    \\&~~~~+
    N (-4p + 20p^2 - 32p^3 + 16p^4) +       
    2 p - 8 p^2 + 12 p^3 - 6 p^4
    \\
    \mathbb{E}[yw] &= 
    2 (N - 1)^2 (N - 2) p^3 (1-p) + 
    2 (N - 1)^2 p^2 (1-p)           
    \\&~~~~+
    2 (N - 1)^2 (N - 2) p (1-p)^3 + 
    2 (N - 1)^2 p (1-p)^3           
    \\&~~~~+
    (N - 1) (N - 2) p (1-p)^3 +   
    (N - 1) p (1-p)^2             
    \\&~~~~+
    (N - 1)^2 p (1-p)^2 +         
    (N - 1) p (1-p)^2             
    \\
    &=
    N^3 (2 p - 6 p^2 + 8 p^3 - 4 p^4) +
    N^2 (-4 p + 15 p^2 - 24 p^3 + 13 p^4) 
    \\&~~~~+
    N (3 p - 13 p^2 + 23 p^3 - 13 p^4) +
    (-p + 4 p^2 - 7 p^3 + 4 p^4)
\end{align*}

Hence,
\begin{align*}
    d_{31} := \mathbb{E}[xy] - \mathbb{E}[y^2] &= -2 (-1 + N) (2 + 2 N (-1 + p) - 3 p) (-1 + p) p^2
    \\
    d_{32} := \mathbb{E}[xz] - \mathbb{E}[yz] &= -2 (-1 + N)^2 (2 + 2 N (-1 + p) - 3 p) (-1 + p) p^2
    \\
    d_{33} := \mathbb{E}[xw] - \mathbb{E}[yw] &= (-1 + p) (p (-1 + 2 p - 4 p^2) + 2 N^3 p (1 - 2 p + 2 p^2) 
    \\&~~~~+
   N p (3 - 8 p + 13 p^2) - N^2 (1 + 2 p - 9 p^2 + 13 p^3))
\end{align*}

\textbf{Conclusion.}

Recall that
$a,b,c > 0$ for any $n \geq 12$ and $0 < \tau_1 \leq 1$
and we aim to show that
\[
\sum_{i,j} a (\mathbb{E}[xy] - \mathbb{E}[y^2]) + b(\mathbb{E}[xz] - \mathbb{E}[yz]) + c(\mathbb{E}[xw] - \mathbb{E}[yw]) < 0.
\]

It suffices to show that, for any $0.5 < p \leq 1$,
\begin{align*}
    \sum_{i,j} (\mathbb{E}[xy] - \mathbb{E}[y^2]) \leq 0
    \sum_{i,j} (\mathbb{E}[xz] - \mathbb{E}[yz]) \leq 0
    \sum_{i,j} (\mathbb{E}[xw] - \mathbb{E}[yw]) \leq 0    
\end{align*}
with at least one inequality being strict.
Since there are 
\begin{itemize}[leftmargin=*]
    \item $2N$ pairs of $(i, j)$ such that $i = j$,
    \item $2N(N-1)$ pairs of $(i, j)$ such that $i \neq j$ and $i$ and $j$ are in the same group, and
    \item $2N^2$ pairs of $(i, j)$ such that $i \neq j$ and $i$ and $j$ are in different groups,
\end{itemize}
it is equivalent to showing that
\[
d_{1i} + (N - 1) d_{2i} + N d_{3i} \leq 0, \forall i \in \{1,2,3\},
\]
and at least for one $i$ value, the inequality is strict.

Indeed,
\begin{align*}
    d_{11} + (N - 1) d_{21} + N d_{31} &=
    -N^4 (1 - 2 p + 2 p^2)^2 
    + 2 N p (-1 + 6 p - 12 p^2 + 10 p^3) 
    \\&~~~~+
    N^2 (-1 + 4 p - 17 p^2 + 29 p^3 - 26 p^4) 
    + N^3 (1 - 6 p + 17 p^2 - 22 p^3 + 16 p^4)
    \\&~~~~+
    (p - 4 p^2 + 9 p^3 - 6 p^4) < 0
\end{align*}
for any $N \geq 4$;
\begin{align*}
    d_{12} + (N - 1) d_{22} + N d_{32} &=
    N^4 (-1 + 4 p - 12 p^2 + 16 p^3 - 8 p^4) +
    N^3 p (-4 + 31 p - 56 p^2 + 34 p^3) 
    \\&~~~~+
    N^2 (p - 33 p^2 + 77 p^3 - 52 p^4) + 
    N p (-1 + 22 p - 52 p^2 + 32 p^3) 
    \\&~~~~+
    p - 8 p^2 + 15 p^3 - 6 p^4 < 0
\end{align*}
for any $N \geq 3$;
\begin{align*}
    d_{13} + (N - 1) d_{23} + N d_{33} &=
    N (-1 + p) (p - 12 p^3 + 4 N^3 p (1 - 2 p + 2 p^2) 
    \\&~~~~+
   N p (1 - 12 p + 33 p^2) - N^2 (1 + 4 p - 19 p^2 + 29 p^3)) < 0,
\end{align*}
for any $N \geq 2$.

In conclusion, when trained with any $0.5 < p_{train} \leq 1$ (i.e., $0 < \tau_1 \leq 1$) and $N_{train} \geq 6$ (i.e., $n \geq 12$), as $\gamma$ approaches $0^+$, for any $0.5 < p_{test} \leq 1$ (i.e., $0 < \tau_2 \leq 1$),
\begin{align*}
    d_1 := d_{11} + (N - 1) d_{21} + N d_{31} &< 0 \\
    d_2 := d_{12} + (N - 1) d_{22} + N d_{32} &< 0 \\
    d_3 := d_{13} + (N - 1) d_{23} + N d_{33} &\leq 0,
\end{align*}
completing the proof.

\end{proof}

\subsection{Proof of Theorem~\ref{theorem:decrease}}

\begin{reptheorem}{theorem:decrease}[Correlation with $\tau$]
    For any $n \geq 34$ and $0 < \tau_1 \leq 0.98$, there exists $\epsilon > 0$
    such that for any $0 < \gamma \leq \epsilon$ and any $\tau_1 + 0.02 \leq \tau_2 \leq 1$ with the initial weight $\mathbf{W}^{(0)} = \mathbb{I}_{n}$,
    \begin{equation*}
       \underbrace{\mathbb{E}_{\mathcal{G}} \left[\mathcal{L}(\mathcal{G},\mathbf{W}^{(0)}) -  \mathcal{L}(\mathcal{G},\mathbb{W}(\tau_{1}, \mathbf{W}^{(0)},\gamma))~\vert~\tau_{1}\right]}_{\text{the decrease in error on graphs of strength $\tau_1$ (training graphs)}} <  \underbrace{\mathbb{E}_{\mathcal{G}}\left[\mathcal{L}(\mathcal{G},\mathbf{W}^{(0)}) -  \mathcal{L}(\mathcal{G},\mathbb{W}(\tau_{1}, \mathbf{W}^{(0)},\gamma))~\vert~\tau_{2}\right]}_{\text{the decrease in error on graphs of strength $\tau_2$}}
    \end{equation*}
\end{reptheorem}
\begin{proof}
    Following the notations in the proof of Theorem~\ref{theorem:generalization} above,
    regarding the ``speed'' of gradient,
\begin{align*}
    \frac{\partial d_1}{\partial p} &= 
    4 N^4 (1 - 4 p + 6 p^2 - 4 p^3) + 
    N^3 (-6 + 34 p - 66 p^2 + 64 p^3) 
    \\&~~~~+
    N^2 (4 - 34 p + 87 p^2 - 104 p^3) + 
    N (-2 + 24 p - 72 p^2 + 80 p^3) 
    \\&~~~~+
    (1 - 8 p + 27 p^2 - 24 p^3) \\
    \frac{\partial d_2}{\partial p} &=
    4 N^4 (1 - 2 p)^3 +  
    2 N^3 (-2 + 31 p - 84 p^2 + 68 p^3) 
    \\&~~~~+
    N^2 (1 - 66 p + 231 p^2 - 208 p^3) + 
    N (-1 + 44 p - 156 p^2 + 128 p^3) 
    \\&~~~~+
    (1 - 16 p + 45 p^2 - 24 p^3) \\
    \frac{\partial d_3}{\partial p} &=    
    4 N^4 (-1 + 2 p)^3 + 
    N^3 (3 - 46 p + 144 p^2 - 116 p^3) 
    \\&~~~~+
    N^2 (-1 + 26 p - 135 p^2 + 132 p^3) +
    N(-1 + 2 p + 36 p^2 - 48 p^3).
\end{align*}

Trained with $p' = \frac{\tau_1 + 1}{2}$ and testing with $p = \frac{\tau_2 + 1}{2}$,
we have
\begin{align*}
    f(p, p', N) &:=\frac{\partial (a d_1 + b d_2 + cd_2)}{\partial p} 
    \\&= \underbrace{c_0(N)}_{\text{a positive constant determined by $N$}} \lim_{\gamma \to 0^+} -\frac{1}{\gamma} \mathbb{E}_{\mathcal{G}}\left[\mathcal{L}(\mathcal{G},\mathbf{W}^{(0)}) -  \mathcal{L}(\mathcal{G},\mathbb{W}(\tau_{1}, \mathbf{W}^{(0)},\gamma))~\vert~\tau_{2}\right]
    \\
    &= a(N, p') \frac{\partial d_1(N, p)}{\partial p} + 
    b(N, p') \frac{\partial d_2(N, p)}{\partial p} + 
    c(N, p') \frac{\partial d_3(N, p)}{\partial p}     
    \\
    &=
    -((-1 + N) N (-1 + 2 p + 36 p^2 - 48 p^3 + 4 N^3 (-1 + 2 p)^3 + 
      N^2 (3 - 46 p + 144 p^2 - 116 p^3) 
      \\&~~~~+
      N (-1 + 26 p - 135 p^2 + 132 p^3)) (-1 + p') (1 + 14 p' - 
      26 p'^2 + 32 p'^3 + 4 N^2 p' (1 - 2 p' + 2 p'^2) 
      \\&~~~~-
      N (1 + 14 p' - 28 p'^2 + 32 p'^3))) - (-1 + 8 p - 27 p^2 + 24 p^3 +
     N (2 - 24 p + 72 p^2 - 80 p^3) 
     \\&~~~~+
    N^3 (6 - 34 p + 66 p^2 - 64 p^3) + 
    4 N^4 (-1 + 4 p - 6 p^2 + 4 p^3) 
    \\&~~~~+
    N^2 (-4 + 34 p - 87 p^2 + 104 p^3)) (18 p'^4 + 
    N^2 (1 + 4 p' - 10 p'^2 + 4 p'^3 + 4 p'^4) 
    \\&~~~~-
    N (1 + 8 p' - 18 p'^2 + 8 p'^3 + 16 p'^4)) - (-1 + 16 p - 45 p^2 + 
    24 p^3 + 4 N^4 (-1 + 2 p)^3 
    \\&~~~~+
    N (1 - 44 p + 156 p^2 - 128 p^3) 
    \\&~~~~-
    2 N^3 (-2 + 31 p - 84 p^2 + 68 p^3) + 
    N^2 (-1 + 66 p - 231 p^2 + 208 p^3)) (-6 p'^3 (1 + 4 p') 
    \\&~~~~+
    N^2 (-3 + 9 p' - 34 p'^2 + 52 p'^3 - 34 p'^4) + 
    N^3 (1 - 4 p' + 12 p'^2 - 16 p'^3 + 8 p'^4) 
    \\&~~~~+
    N (2 - 3 p' + 22 p'^2 - 38 p'^3 + 48 p'^4)).
\end{align*}

Given $p_1 = \frac{\tau_1 + 2}{2}$ and $p_2 = \frac{\tau_2 + 2}{2}$,
\begin{align*}
    &~f(p_2, p_1, N) - f(p_1, p_1, N) \\
    &= 
    (-1 + N) N (-1 + p_1) (1 + 14 p_1 - 26 p_1^2 + 32 p_1^3 + 
    4 N^2 p_1 (1 - 2 p_1 + 2 p_1^2) 
    \\&~~~~-
    N (1 + 14 p_1 - 28 p_1^2 + 32 p_1^3)) (-1 + 2 p_1 + 36 p_1^2 - 
    48 p_1^3 + 4 N^3 (-1 + 2 p_1)^3 
    \\&~~~~+
    N^2 (3 - 46 p_1 + 144 p_1^2 - 116 p_1^3) + 
    N (-1 + 26 p_1 - 135 p_1^2 + 132 p_1^3)) 
    \\&~~~~+
    (-1 + 8 p_1 - 27 p_1^2 + 
    24 p_1^3 + N (2 - 24 p_1 + 72 p_1^2 - 80 p_1^3) + 
    N^3 (6 - 34 p_1 + 66 p_1^2 - 64 p_1^3) 
    \\&~~~~+
    4 N^4 (-1 + 4 p_1 - 6 p_1^2 + 4 p_1^3) 
    \\&~~~~+
    N^2 (-4 + 34 p_1 - 87 p_1^2 + 104 p_1^3)) (18 p_1^4 + 
    N^2 (1 + 4 p_1 - 10 p_1^2 + 4 p_1^3 + 4 p_1^4) 
    \\&~~~~-
    N (1 + 8 p_1 - 18 p_1^2 + 8 p_1^3 + 16 p_1^4)) + (-1 + 16 p_1 - 
    45 p_1^2 + 24 p_1^3 + 4 N^4 (-1 + 2 p_1)^3 
    \\&~~~~+
    N (1 - 44 p_1 + 156 p_1^2 - 128 p_1^3) - 
    2 N^3 (-2 + 31 p_1 - 84 p_1^2 + 68 p_1^3) 
    \\&~~~~+
    N^2 (-1 + 66 p_1 - 231 p_1^2 + 208 p_1^3)) (-6 p_1^3 (1 + 4 p_1) + 
    N^2 (-3 + 9 p_1 - 34 p_1^2 + 52 p_1^3 - 34 p_1^4) 
    \\&~~~~+
    N^3 (1 - 4 p_1 + 12 p_1^2 - 16 p_1^3 + 8 p_1^4) + 
    N (2 - 3 p_1 + 22 p_1^2 - 38 p_1^3 + 48 p_1^4)) 
    \\&~~~~-
    (18 p_1^4 + 
    N^2 (1 + 4 p_1 - 10 p_1^2 + 4 p_1^3 + 4 p_1^4) 
    \\&~~~~-
    N (1 + 8 p_1 - 18 p_1^2 + 8 p_1^3 + 16 p_1^4)) (-1 + 8 p_2 - 27 p_2^2 + 
    24 p_2^3 + N (2 - 24 p_2 + 72 p_2^2 - 80 p_2^3) 
    \\&~~~~+
    N^3 (6 - 34 p_2 + 66 p_2^2 - 64 p_2^3) + 
    4 N^4 (-1 + 4 p_2 - 6 p_2^2 + 4 p_2^3) 
    \\&~~~~+
    N^2 (-4 + 34 p_2 - 87 p_2^2 + 104 p_2^3)) - (-1 + N) N (-1 + 
    p_1) (1 + 14 p_1 - 26 p_1^2 + 32 p_1^3 
    \\&~~~~+
    4 N^2 p_1 (1 - 2 p_1 + 2 p_1^2) - 
    N (1 + 14 p_1 - 28 p_1^2 + 32 p_1^3)) (-1 + 2 p_2 + 36 p_2^2 
    \\&~~~~-
    48 p_2^3 + 4 N^3 (-1 + 2 p_2)^3 
    \\&~~~~+
    N^2 (3 - 46 p_2 + 144 p_2^2 - 116 p_2^3) + 
    N (-1 + 26 p_2 - 135 p_2^2 + 132 p_2^3)) - (-6 p_1^3 (1 + 4 p_1) 
    \\&~~~~+
    N^2 (-3 + 9 p_1 - 34 p_1^2 + 52 p_1^3 - 34 p_1^4) 
    \\&~~~~+
    N^3 (1 - 4 p_1 + 12 p_1^2 - 16 p_1^3 + 8 p_1^4) + 
    N (2 - 3 p_1 + 22 p_1^2 - 38 p_1^3 + 48 p_1^4)) (-1 + 16 p_2 
    \\&~~~~-
    45 p_2^2 + 24 p_2^3 + 4 N^4 (-1 + 2 p_2)^3 
    \\&~~~~+
    N (1 - 44 p_2 + 156 p_2^2 - 128 p_2^3) - 
    2 N^3 (-2 + 31 p_2 - 84 p_2^2 + 68 p_2^3) 
    \\&~~~~+
    N^2 (-1 + 66 p_2 - 231 p_2^2 + 208 p_2^3))
\end{align*}

By analyzing the above equation, we can find that 
when $N \geq 17$ (i.e., $n \geq 34$), $0.5 < p_1 \leq 0.99$ (i.e., $0 < \tau_1 \leq 0.98$), and $p_1 + 0.01 \leq p_2 \leq 1$ (i.e., $\tau_1 + 0.02 \leq \tau_2 \leq 1$),
the above equation is negative, i.e., $f(p_2, p_1, N) - f(p_1, p_1, N) < 0$ or equivalently
\begin{align*}   
\lim_{\gamma \to 0^+} \frac{1}{\gamma} (&\mathbb{E}_{\mathcal{G}}\left[\mathcal{L}(\mathcal{G},\mathbf{W}^{(0)}) - \mathcal{L}(\mathcal{G},\mathbb{W}(\tau_{1}, \mathbf{W}^{(0)},\gamma))~\vert~\tau_{1}\right] \\ 
- 
&\mathbb{E}_{\mathcal{G}}\left[\mathcal{L}(\mathcal{G},\mathbf{W}^{(0)}) -  \mathcal{L}(\mathcal{G},\mathbb{W}(\tau_{1}, \mathbf{W}^{(0)},\gamma))~\vert~\tau_{2}\right]) < 0,
\end{align*}
completing the proof.
\end{proof}

\section{Datasets}
\label{sec:datasets}
In this section, we describe the details of the benchmark graph classification datasets utilized in our experiments.
For experiments, we use datasets that are preprocessed by~\citet{Morris+2020}.
We provide descriptive statistics of each dataset in Table~\ref{tab:datasets}.

The \textbf{DD} dataset~\citep{dobson2003distinguishing} is a bioinformatics dataset that contains protein graphs. 
Each graph represents a particular protein.
Each node represents an individual amino acid.
Each edge represents two nodes that satisfy certain spatial proximity.
Each node feature represents a type of the corresponding node's amino acid type.
Each graph binary label represents whether the corresponding compound is an enzyme or not.

The \textbf{{Protein}} dataset~\citep{borgwardt2005protein} is a bioinformatics dataset that contains protein graphs. 
Each graph represents a particular protein.
Each node represents a secondary structure element of a protein.
Each edge represents two nodes that are neighbors along the amino acid sequence or are top-K neighbors (K=3) at the protein space.
Each node feature represents a type of the corresponding node's secondary structure element type.
Each graph binary label represents whether the corresponding protein is an enzyme or not.

The \textbf{NCI1} dataset~\citep{wale2008comparison} is a chemical dataset that contains chemical molecular compound graphs.
Each graph represents a particular chemical molecular compound.
Each node represents an atom.
Each edge represents a chemical bond between two atoms.
Each node feature represents a type of atom (e.g., carbon, and hydrogen).
Each graph binary label represents whether the corresponding compound is an anti-cancer chemical or not.

The \textbf{AIDS} dataset~\citep{riesen2008iam} is a chemical dataset that contains chemical molecular compound graphs.
Each graph represents a particular chemical molecular compound.
Each node represents an atom.
Each edge represents a chemical bond between two atoms.
Each node feature represents a type of atom (e.g., carbon, and hydrogen).
Each graph binary label represents whether the corresponding compound shows an anti-HIV activity or not.

The \textbf{Reddit} dataset~\citep{yanardag2015deep} is a social-network dataset that contains social network graphs.
Among Reddit-Binary, Reddit-Multi-5K, and Reddit-Multi-12K, we utilized Reddit-Binary.
Each graph represents a thread on Reddit.
Each node represents a user.
Each edge represents a comment between two users.
There are no available node features, thus, we utilize the one-hot encoded degree of the node as its feature, following existing literature in GLAD~\citep{qiu2022raising}.
Each graph binary label represents whether the corresponding thread is a discussion-based thread or question-answer-based thread.

The \textbf{IMDB} dataset~\citep{yanardag2015deep} is an actor-ego-network dataset that contains social network graphs.
Each graph represents the ego network of a particular actor.
Each node represents an actor.
Each edge represents a co-appearance of two actors in (a) movies.
There are no available node features, thus, we utilize the one-hot encoded degree of the node as its feature, following existing literature in GLAD~\citep{qiu2022raising}.
Each graph binary label represents whether the corresponding ego-user particularly acts in the romance genre or action genre.

\begin{table}[t]
\caption{Descriptive statistics of the utilized graph classification benchmark datasets.}
\setlength{\tabcolsep}{5pt}
\small
\centering
\scalebox{1.0}{
\renewcommand{\arraystretch}{1.2}
\begin{tabular}{l|c c c c c c c c c c c c}
    \toprule
        Methods & DD & Protein & NCI1 & AIDS & Reddit & IMDB & MUTAG & DHFR & BZR & ER\\
    \midrule 
    \midrule 
     \# of graphs & 1,178 & 1,113 & 4,110 & 2,000 & 2,000 & 1,000 & 4,337 & 756 & 405 & 7,697 \\

     Avg. $\vert \mathcal{V} \vert$ & 284.32 & 39.06 & 29.87 & 15.69 & 429.63 & 19.77 & 30.32 & 42.43 & 35.75  & 17.58 \\
     
     Avg. $\vert \mathcal{E} \vert$ & 715.66 & 72.82 & 32.30 & 16.20 & 497.75 & 96.53 & 30.77 & 44.54 & 38.36 & 17.94 \\

     \# of features & 89 & 4 & 37 & 42 & - & - & 14 & 56 & 56 & 50 \\

     \# of classes & 2 & 2 & 2 & 2 & 2 & 2 & 2 & 2 & 2 & 2 \\

     \bottomrule
\end{tabular}
\label{tab:datasets}
}
\end{table}

The \textbf{MUTAG} (formally, mutagenicity) dataset~\citep{kazius2005derivation} is a chemical dataset that contains chemical molecular compound graphs.
Each graph represents a particular chemical molecular compound.
Each node represents an atom.
Each edge represents a chemical bond between two atoms.
Each node feature represents a type of atom (e.g., carbon, and hydrogen).
Each graph binary label represents whether the corresponding compound is mutagenic or not. 

The \textbf{DHFR} dataset~\citep{sutherland2003spline} is a chemical dataset that contains chemical molecular compound graphs.
Each graph represents a particular chemical molecular compound.
Each node represents an atom.
Each edge represents a chemical bond between two atoms.
Each node feature represents a type of atom (e.g., carbon, and hydrogen).
Each graph binary label represents whether the corresponding compound works as a dihydrofolate reductase inhibitor or not.

The \textbf{BZR}~\citep{sutherland2003spline} dataset is a chemical dataset that contains chemical molecular compound graphs.
Each graph represents a particular chemical molecular compound.
Each node represents an atom.
Each edge represents a chemical bond between two atoms. 
Each node feature represents a type of atom (e.g., carbon, and hydrogen).
Each graph binary label represents whether the corresponding compound is active against the benzodiazepine receptor.

The \textbf{ER} dataset is a chemical dataset that contains chemical molecular compound graphs.
Specifically, it contains compounds that are NR-estrogen receptor (ER)-LBD.
Each graph represents a particular chemical molecular compound.
Each node represents an atom.
Each edge represents a chemical bond between two atoms. 
Each node feature represents a type of atom (e.g., carbon, and hydrogen).
Each graph binary label represents whether the corresponding compound has toxicity or not.

The source of the ER dataset is~\url{https://tripod.nih.gov/tox21/challenge/data.jsp}.

\section{Formal Expression of Synthetic Datasets}
\label{sec:syntheticformal}
In this section, we formally describe the leveraged synthetic graphs: Syn-Com and Syn-Cycle.
We elaborate on how each graph is generated.

\subsection{Syn-Com} 
We formalize the data generation process of {Syn-Com}. 
For a given even number of nodes $n > 2$, assume two disjoint node communities $\mathcal{V}_{a}$ and $\mathcal{V}_{b}$ (WLOG, $\mathcal{V}_{a} = \{v_{i} : 1 \leq i \leq n/2\}$ and $\mathcal{V}_{b} = \{v_{j} : n/2 + 1 \leq j \leq n\}$). 
For a given community parameter $\tau \in [0, 1]$, a graph of Syn-Com $\mathcal{G} = (\mathbf{X}, \mathbf{A})$ satisfies the following: $\mathbf{A}_{i,j} = \mathbf{A}_{j,i} \sim b(1, (\tau+1)/2)$ for $\{v_{i},v_{j}\} \in \binom{\mathcal{V}_{a}}{2} \cup \binom{\mathcal{V}_{b}}{2}$, $\mathbf{A}_{i,j}=\mathbf{A}_{j,i} \sim b(1,(1-\tau)/2)$ for $v_{i} \in \mathcal{V}_{a}, v_{j} \in \mathcal{V}_{b}$, and $\mathbf{A}_{i,i} = 0, \forall i\in [n]$, where $b(1,p)$ is a Bernoulli sampling with a parameter $p$. 
We let $\mathbf{X} = \mathbb{I}_{n}$, where $\mathbb{I}_{n}$ is a $n$-by-$n$ identity matrix.

\subsection{Syn-Cycle} 
We formalize the data generation process of Syn-Cycle. 
For a given number of nodes $n > 2$, we consider a clean-cycle graph $\mathcal{G}^{cyc}_{*}=(\mathcal{V},\mathcal{E}^{*})\equiv (\mathbf{X}, \mathbf{A}^{*})$ where $\mathcal{E}^{*} = \{v_{i},v_{i+1} : i \in [n]\} \cup \{v_{1}, v_{n}\}$ and $\mathbf{A}^{*}$ is the adjacency matrix that represents $\mathcal{E}^{*}$.

We also consider a set of noisy-cycle graphs (i.e., pan graphs) $\mathbb{G}^{cyc}_{'}$, constructed by randomly relocating an edge from $\mathcal{G}^{cyc}_{*}$ to form a ($n-1$)-size cycle,
(i.e., $(\mathbf{X}, \{v_{i}, v_{j\vert i}\} \cup \mathcal{E}^{*}\setminus \{e_{k}\}), (\mathbf{X}, \{v_{j}, v_{i\vert j}\} \cup \mathcal{E}^{*}\setminus \{e_{k}\}) \in \mathbb{G}^{cyc}_{'},\forall e_{k} = \{v_{i},v_{j}\} \in \mathcal{E}^{*}$, where $v_{j\vert i}$ is a $v_{j}$'s neighbor except for $v_{i}$).
We let $\mathbf{X} = \mathbb{I}_{n}$.
Note that for $n$ nodes, there exist $2n$ noisy-cycle graphs in $\mathbb{G}^{cyc}_{'}$ and a single clean-cycle graph.

\section{Experimental Details}
\label{sec:experimentdetails}
In this section, we provide experimental details of the conducted experiments in our work.

\subsection{Machines}\label{subsec:machines}
All experiments of this work are performed on a machine with NVIDIA RTX 8000 D6 GPUs (48GB VRAM)
and two Intel Xeon Silver 4214R processors.

\subsection{Details of empirical analysis}\label{subsec:analysisdetails}

\textbf{For the Com-Com dataset}, we sample $500$ graphs from $\mathbb{G}^{com}_{\tau = 0.4}$ for training graphs and sample $500$ graphs from $\mathbb{G}^{com}_{\tau = 0.8}$ for unseen graphs.

\textbf{For the Cycle-Cycle dataset}, we sample half of $\mathbb{G}_{'}^{cyc}$ for training graphs and employ the clean-cycle graph as the unseen graph.

\textbf{For the Com-Cycle dataset}, we sample $500$ graphs from $\mathbb{G}^{com}_{\tau = 0.4}$ for training graphs and sample half of $\mathbb{G}_{'}^{cyc}$ for unseen graphs.

\textbf{For the Cycle-Com dataset}, we sample half of $\mathbb{G}_{'}^{cyc}$ for training graphs and sample 500 graphs from $\mathbb{G}^{com}_{\tau = 0.4}$ for unseen graphs.

\textbf{Training and evaluation} 
We train GAEs for 200 epochs. The results are similar for 500 epochs, but we present the 200-epoch results for better visualization since the error lines flatten after 200 epochs.
For every 10 epochs of training, we measure the mean reconstruction errors for each graph and report the mean of these values for graphs within the same class.

\subsection{Details of two-stage GLAD methods}\label{subsec:twostagedetails}

We provide an overview of two-stage anomaly detection methods.
First, by performing certain pretext tasks, they learn representations of data.
Then, upon the obtained representations, they utilize a one-class classifier to detect anomalies.
Below, (1) we describe graph representations of utilized GLAD methods, including \method, and (2) describe the utilized one-class classifier, an MLP autoencoder.

\textbf{\textit{Note}}. \method \textit{is a two-stage method.}

\textbf{Graph representation.}
As described in Section~\ref{sec:method}, 
 \method obtains a representation of each graph with the proposed error representation Eq~(\ref{eq:mainerrorvector}).
On the other hand, utilized baseline two-stage approaches, which are GraphCL~\citep{you2020graph}, GAE~\citep{kipf2016semi}, and GraphMAE~\citep{hou2022graphmae}, obtain graph representations from the graph or node embeddings.
Specifically, they first train graph neural network encoders with their pretext tasks. 
Specifically, GraphCL performs graph contrastive learning, GAE performs adjacency matrix reconstruction, and GraphMAE performs node feature reconstruction.
After finalizing self-supervised learning, we utilize the trained encoder to obtain embeddings of graphs or nodes.
Details of each method are as follows:
\begin{itemize}[leftmargin = *]
    \item GraphCL: It learns graph embeddings to perform its contrastive learning task. 
    Thus, we directly utilize the graph embeddings as representations of graphs.
    \item GAE: It only learns node embeddings. 
    Thus, we pool node embeddings by using the elementwise-mean readout function.
    \item GraphMAE: It only learns node embeddings. 
    Thus, we pool node embeddings by using the elementwise-mean readout function.
\end{itemize}

\textbf{One-class classifier.}
In our experiments, every two-stage method leverages an MLP autoencoder as its one-class classifier.
In a nutshell, the MLP autoencoder detects anomalies by using the reconstruction loss of a graph.
We first elaborate on how we train the MLP autoencoder and then describe how the trained MLP autoencoder is leveraged to detect anomalies.

We start with the training of the MLP autoencoder.
Formally, for a given data representation vector $\mathbf{z} \in \mathbb{R}^{d}$, the MLP autoencoder $\texttt{MLP}_{\xi} : \mathbb{R}^{d} \mapsto \mathbb{R}^{d}$ generates a reconstructed representation $\hat{\mathbf{z}} \in \mathbb{R}^{d}$ (i.e., $\hat{\mathbf{z}} = \texttt{MLP}_{\xi}(\mathbf{z})$).
Then, we compute a L2-norm reconstruction loss $\mathcal{L}$ as follows: $\mathcal{L} \coloneqq \lVert \mathbf{z} - \hat{\mathbf{z}} \rVert_{2}$.
We update the parameters of the MLP autoencoder $\xi$ by using gradient descent to minimize $\mathcal{L}$. 

{After training the MLP autoencoder $\texttt{MLP}_{\xi}$, we detect anomalies by using a reconstruction loss as an anomaly score. 
However, since the scale of each representation dimension significantly varies, the reconstruction loss is often dominated by dimensions with larger scales, overshadowing those with smaller values.
To mitigate this undesirable phenomenon, in the anomaly inference step, we leverage dimensionwise-weighted L2 scores. 
Importantly, we apply different weights to different dimensions $w_{l} \in \mathbb{R},\forall l \in [d]$ (note that $d$ is the dimension of an input vector).
Specifically, we utilize the dimensionwise standard deviation of train data's representations, which is defined as follows:
\begin{equation}\label{eq:columnweight}
    w_{\ell} = \sqrt{\frac{1}{\vert \mathcal{D}_{train}\vert}\sum_{\mathcal{G}_{k} \in \mathcal{D}_{train}}\left(\mathbf{z}_{k,\ell} - \left(\frac{1}{\vert \mathcal{D}_{train}\vert} \sum_{\mathcal{G}_{t} \in \mathcal{D}_{train}} \mathbf{z}_{t,\ell}\right)\right)^{2}}, \forall \ell \in [d].
\end{equation}
where $\mathbf{z}_{i,l}$ is a $l$-th element of $i$-th graph's representation vector and $\mathcal{D}_{train}$ is a set of training graphs. 
By using the normalization weight of Eq~(\ref{eq:columnweight}), we compute the anomaly score. 
Formally, the anomaly detection score of $\mathcal{G}_{t}$, denoted by $s_{T}$, is defined as follows:
\begin{equation}
 s_t \coloneqq \exp\left(-\sqrt{\sum_{\ell = 1}^{d}\left( \frac{\mathbf{z}_{t, \ell}' - \texttt{MLP}_{\xi}(\mathbf{z}_{t}')_{\ell}}{w_{\ell}}  \right)^{2}}\right)\in [0, 1].
\end{equation}}
Note that the higher $s_{t}$ indicates the $t$-th data point is normal.

\subsection{Details of \method}\label{subsec:aggregationdetail}

In this subsection, we provide details of \method.

\textbf{Reconstruction models.}
As described in Section~\ref{sec:experiment}, we use GIN~\citep{xu2018powerful} as our backbone encoder, which is the same as other leveraged GLAD baseline methods.
For the node feature decoder and adjacency matrix decoder, we use a 2-layer MLP with a ReLU activation function.

\textbf{Error representation.}
For aggregation functions of error representation, we leverage a mean function and a standard deviation function. 
Formally, for a set $\mathcal{A}$ where $\mathcal{A} \subset \mathbb{R}$, a mean aggregation function $\texttt{Agg}_{\texttt{mean}} : 2^{\mathbb{R}} \mapsto \mathbb{R}$ and a standard deviation aggregation function $\texttt{Agg}_{\texttt{std}} : 2^{\mathbb{R}} \mapsto \mathbb{R}_{\geq 0}$ are defined as follows:
\begin{equation}
    \texttt{Agg}_{\texttt{mean}}(\mathcal{A}) \coloneqq \frac{1}{\vert \mathcal{A}\vert} \sum_{a\in \mathcal{A}} a, \quad 
    \texttt{Agg}_{\texttt{std}}(\mathcal{A}) \coloneqq \sqrt{\frac{1}{\vert \mathcal{A}\vert} \sum_{a \in \mathcal{A}}(a - \texttt{Agg}_{\texttt{mean}}(\mathcal{A}))^{2}}.
\end{equation}
Thus, we represent each graph $\mathcal{G}$ by using a $\mathbb{R}^{4}$ vector as follows:
\begin{equation}
    \texttt{Err}(\mathcal{G}) = [\texttt{Agg}_{\texttt{mean}}(\mathbb{L}_{\mathbf{X}}(\mathcal{G})),~ 
    \texttt{Agg}_{\texttt{std}}(\mathbb{L}_{\mathbf{X}}(\mathcal{G})),~
    \texttt{Agg}_{\texttt{mean}}(\mathbb{L}_{\mathbf{A}}(\mathcal{G})),~
    \texttt{Agg}_{\texttt{std}}(\mathbb{L}_{\mathbf{A}}(\mathcal{G}))] \in \mathbb{R}^{4}.
\end{equation}

\subsection{Hyperparameters and Details of Baselines}

For every method, we tune the hyperparameters and choose the hyperparameter configuration that gives the best validation AUROC.
We provide details of the search space of each model category and its hyperparameters.

\textbf{Fixed settings.} 
For all the methods, we fix the dropout probability and weight decay as $0.3$ and $1e-6$, respectively.
In addition, all the methods are trained with Adam optimizer~\citep{kingma2014adam}.

\textbf{One-stage learning-based GLAD methods.} 
We tune the following hyperparameters for one-stage learning-based GLAD methods, which are DOMINANT, OCGTL, GLADC, GLocalKD, GLAM, HIMNET, and SIGNET:
\begin{itemize}[leftmargin=*]
    \item Training learning rate $\gamma \in \{10^{-3}, 10^{-4}\}$
    \item Models hidden dimension $d' \in \{16, 32, 64, 128, 256\}$
    \item Number of GNN layers $K \in \{3,4,5\}$
    \item Number of model training epochs $L \in \{30, 60, \cdots 300\}$
\end{itemize}
For other hyperparameters, we follow the default configuration provided in their official Github.

\textbf{Two-stage learning-based GLAD methods.} 
We tune the following hyperparameters for two-stage learning-based GLAD methods, which are GraphCL, GAE, GraphMAE, and \method.

\textbf{\textit{Note.}} Our proposed method \method is also included in this category and tuned hyperparameters according to the below

\begin{itemize}[leftmargin=*]
    \item Pretext task learning rate $\gamma \in \{10^{-3}, 10^{-4}\}$
    \item Models hidden dimension $d' \in \{16, 32, 64, 128, 256\}$
    \item Number of GNN layers $K \in \{3,4,5\}$
    \item Number of pretext task training epochs $L \in \{20, 40, \cdots , 200\}$
\end{itemize}

In addition to these hyperparameters, we also tune the positive weight coefficient for our adjacency matrix reconstruction, which is denoted as $\tau$ in Section~\ref{sec:method}, within $\{0.0, 1.0, 2.0\}$.

Here, models indicate both the encoder model and decoder model that are utilized to perform pretext tasks.
Note that as described in Appendix~\ref{subsec:twostagedetails}, all two-stage methods utilize MLP autoencoder for their one-class classifier. 
We tune the hyperparameter of the MLP autoencoder as follows:
\begin{itemize}[leftmargin=*]
    \item MLP hidden dimension $d' \in \{32, 64, 128\}$
    \item MLP learning rate $\gamma \in \{10^{-2}. 10^{-3}, 10^{-4}\}$
\end{itemize}

We fix the MLP auteoncoder training epochs and the number of MLP layers as $500$ and $3$, respectively.

\textbf{Two-stage baselines details.}
For GraphCL, we obtain graph views by randomly dropping 50\% of edges and adding Gaussian noise to node features sampled from $\mathcal{N}(0,0.1)$ and use a two-layer MLP projection head.
For GraphMAE, we mask $50\%$ of node features with the zero mask, which is empirically demonstrated to be more effective than a learnable mask in our preliminary study.
For GAE, we utilize a binary cross-entropy loss and a two-layer MLP decoder.~\footnote{While the original GAE~\citep{kipf2016variational} does not use a projection head, we found that projection head enhances the performance of GAE, through our preliminary study.}

\section{Additional Experimental Results}
\label{sec:additionalresults}
\begin{figure}[t] 
    \centering
    \vspace{2mm}
    
    \includegraphics[width=\linewidth]{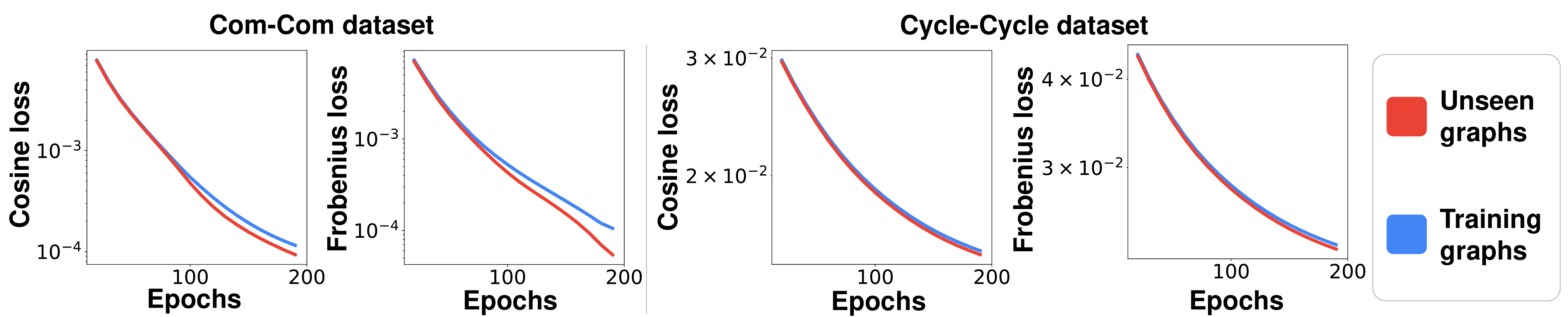}

    \caption{\label{fig:empiricallyrealfeat}
    \textbf{Reconstruction flip occurs on feature reconstruction method.} When Graph-AEs are trained on graphs sharing a primary pattern of weaker strength, the trained Graph-AEs give smaller reconstruction losses for graphs having the same pattern of a stronger strength (\textcolor{googlered}{red} lines) than those of a weaker strength (\textcolor{googleblue}{blue} lines).
    }
\end{figure}

\begin{figure}[t] 
    \centering
    \vspace{2mm}
    
    \includegraphics[width=\linewidth]{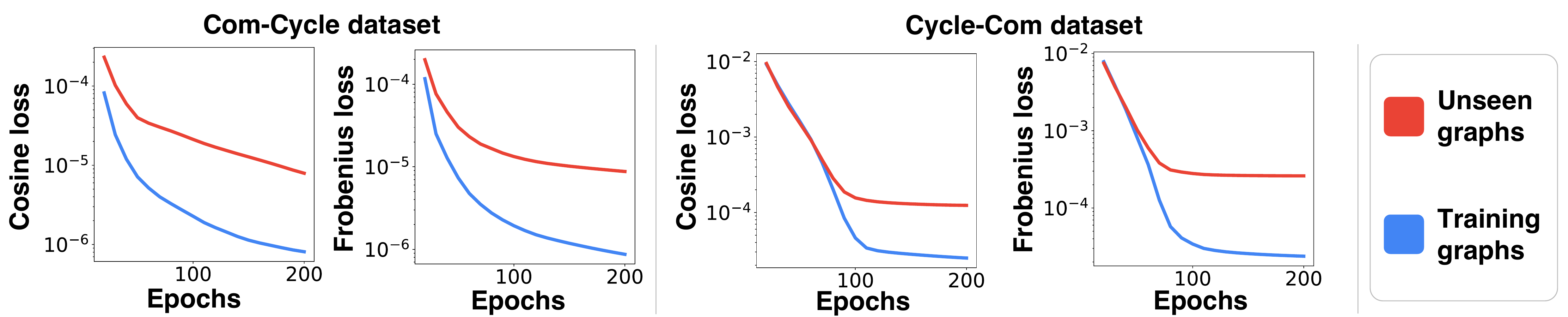}

    \caption{\label{fig:empiricallygoodfeat}
    \textbf{Reconstruction flip does NOT occur on feature reconstruction method.}
    When Graph-AEs are trained on graphs sharing a primary pattern, the trained Graph-AEs give larger reconstruction for graphs having a different pattern (\textcolor{googlered}{red} lines) than those of the same pattern (\textcolor{googleblue}{blue} lines).
    }
     \vspace{2mm}
\end{figure}

\subsection{Node feature reconstruction methods}\label{subsec:nodefeat}
Note that in Section~\ref{sec:analysis}, we investigate Graph-AEs that reconstruct the adjacency matrix of a given graph.
In this subsection, we analyze another large branch of Graph-AEs: node feature reconstruction methods~\citep{hou2022graphmae, hou2023graphmae2,tu2023rare}.

\textbf{Feature reconstruction model.}
We first formalize the node feature reconstruction method.
Consider a graph $\mathcal{G} = (\mathbf{X}, \mathbf{A})$, where $\mathbf{X} \in \mathbb{R}^{\vert \mathcal{V}\vert \times d}$.
First,  a GNN encoder $f_{\theta}$ is utilized to generates node embeddings $\mathbf{Z} \in \mathbb{R}^{\vert \mathcal{V} \vert \times d'}$ (i.e., $f_{\theta}(\mathbf{X}, \mathbf{A})=\mathbf{Z}$).
Then, by using the feature decoder $g_{\phi} : \mathbb{R}^{d'} \mapsto \mathbb{R}^{d}$, the reconstructed node features $\hat{\mathbf{X}} \in \mathbb{R}^{\vert \mathcal{V} \vert \times d}$ are obtained.
Lastly, we compute the feature reconstruction loss, which is either (1) squared Frobenius norm loss~\citep{niu2023graph} (i.e., $\mathcal{L} \coloneqq \lVert \mathbf{X} - \hat{\mathbf{X}}\rVert_{F}^{2}$) or (2) cosine similarity loss~\citep{hou2022graphmae} (i.e., $\mathcal{L} \coloneqq \frac{1}{\vert \mathcal{V}\vert}\sum^{\vert \mathcal{V}\vert}_{i=1}(1 - \frac{\mathbf{X}^{T}_{i}\hat{\mathbf{X}}_{i}}{\lVert \mathbf{X}_{i} \rVert_{2} \cdot \lVert \hat{\mathbf{X}}_{i} \rVert_{2}})$).
Parameters of the GNN encoder $f_{\theta}$ and the node feature decoder $g_{\phi}$ are updated by gradient descent to minimize $\mathcal{L}$.

\textbf{Setting.}
The dataset setting is equal to that of empirical analysis provided in Section~\ref{subsec:empiricalanalysis}.
Regarding the model setting, we utilize 3-layer GIN as the GNN encoder $f_{\theta}$ and $2-$layer MLP as the feature decoder $g_{\phi}$.
We utilize either squared Frobenius norm or cosine similarity loss for the reconstruction loss.
We train each reconstruction model for 200 epochs.

\textbf{Results.}
In a nutshell, we can observe the same results as those of GAE.
In Scenario 1, as shown in Figure~\ref{fig:empiricallyrealfeat}, one can observe that the reconstruction flip occurs, which is also observed in GAE.
In Scenario 2, as shown in Figure~\ref{fig:empiricallygoodfeat}, one can observe that the reconstruction flip does not occur.
Thus, we verify that our findings in Section~\ref{subsec:empiricalanalysis} are not only restricted to the adjacency matrix reconstruction methods but also hold for the node feature reconstruction methods.

\subsection{Experimental results on each setting.}

\begin{table}[t]
\caption{Average and standard deviation of the test AUROC when graphs belonging to class 0 are anomalies and graphs belonging to class 1 are normal.
A.R. denotes the average ranking.
\method outperforms other GLAD methods in terms of average ranking.}
\setlength{\tabcolsep}{2pt}
\small
\centering
\scalebox{0.8}{
\renewcommand{\arraystretch}{1.1}
\begin{tabular}{l|c c c c c c c c c c c c c}
    \toprule
        Methods & DD & Protein & NCI1 & AIDS & Reddit & IMDB & MUTAG & DHFR & BZR & ER & A.R.\\
    \midrule 
    \midrule 
     
     DOMINANT~\citep{ding2019deep} & 44.5 {\std (3.0)} & 43.5 {\std (11.6)} & 66.1 {\std (6.0)} & 71.1 {\std (5.3)} & 34.8 {\std (5.5)} & 59.9 {\std (3.7)} & 59.8 {\std (5.7)} & 49.6 {\std (9.2)} &  72.3 {\std (10.3)} & 66.8 {\std (4.4)} & 12.7\\
     
     OCGTL~\citep{qiu2022raising} & 77.9 {\std (2.6)} &  67.8 {\std (10.6)} & 64.1 \std{(4.0)} & 99.1 {\std (1.5)} & 75.3 {\std (3.5)} & 69.1 {\std (3.3)} & 63.1 {\std (6.6)} &  69.7 {\std (5.9)} & 75.2 {\std (11.3)} & 65.8 {\std (4.7)} & 6.0\\
     
     GLocalKD~\citep{ma2022deep} & 69.6 {\std (7.1)} & 62.5 {\std (3.3)} & 45.9 {\std (6.6)} & 97.5 {\std (0.8)} & 79.0 {\std (2.9)} & 68.1 {\std (5.5)} & 55.8 {\std (2.7)} & 61.8 {\std (9.6)} &  62.8 {\std (12.8)} & 41.0 {\std (5.0)} & 11.2 \\
     
     GLADC~\citep{luo2022deep} & 74.7 {\std (3.6)} & 60.5 {\std (2.7)} & 43.0 {\std (4.5)} & 98.0 {\std (0.4)} & 83.3 {\std (3.2)} & 68.6 {\std (3.5)} & 55.5 {\std (4.0)} & 62.0 {\std (2.9)} &  63.5 {\std (13.0)} & 39.8 {\std (6.3)}  & 11.2\\
     
     GLAM~\citep{zhao2022graph} & 46.6 {\std (5.8)} & 62.6 {\std (3.9)} & 60.7 {\std (1.9)} & 94.2 {\std (3.0)} & 72.2 {\std (4.5)} & 63.7 {\std (2.4)} & 60.6 {\std (1.5)} & 48.3 {\std (6.2)} & 62.1 {\std (8.2)} & 66.6 {\std (2.6)} &12.6\\
     
     HIMNET~\citep{niu2023graph} & 70.2 {\std (2.7)} & 66.0 {\std (2.9)} &  41.1 {\std (5.1)} &  97.9 {\std (0.5)}  & 72.6 {\std (1.9)} & 68.8 {\std (3.8)} & 58.6 {\std (2.4)} & 60.6 {\std (6.9)} & 69.5 {\std (7.1)} & 52.5 {\std (3.4)} & 10.8 \\
     
     SIGNET~\citep{liu2024towards} & 63.7 {\std (9.1)} & 58.1 {\std (6.0)} & 62.8 {\std (5.1)} &  98.1 {\std (2.1)} &  84.8 {\std (4.8)} & 54.3 {\std (4.0)} & 67.1 {\std (1.5)} & 18.1 {\std (7.1)} & 74.6 {\std (6.3)} & 56.5 {\std (3.3)} & 10.3\\
     
     \midrule 

     
     GraphCL-1~\citep{you2020graph} & 49.0 {\std (5.5)} & 51.3 {\std (3.3)} & 66.4 {\std (3.9)} & 44.8 {\std (12.7)} & 58.8 {\std (5.3)} & 45.8 {\std (6.7)} & 53.6 {\std (1.5)} & 51.4 {\std (8.3)} & 56.5 {\std (7.9)} &  63.2 {\std (3.4)} & 15.4\\
     
     GAE-1~\citep{kipf2016variational} & 67.8 {\std (6.1)} & 76.7 {\std (4.9)} & 59.0 {\std (1.8)} & 99.9 {\std (0.1)} & 86.3 {\std (2.1)} & 60.6 {\std (8.2)} & 55.1 {\std (3.1)} & 59.0 {\std (11.6)} & 61.4 {\std (8.8)} & 60.2 {\std (3.5)} & 10.3\\
     
     GraphMAE-1~\citep{hou2022graphmae} & 50.4 {\std (4.9)} & 62.5 {\std (3.5)} & 51.7 {\std (3.7)} & 90.2 {\std (9.5)} & 74.0 {\std (3.0)} & 71.8 {\std (5.9)} & 62.6 {\std (2.4)} & 52.3 {\std (9.2)} & 64.1 {\std (5.2)} & 65.4 {\std (4.6)}  & 11.3 \\
     
     \midrule 
     
    GraphCL-2~\citep{you2020graph} & 59.0 {\std (3.3)} & 51.8 {\std (1.8)} & 62.3 {\std (4.3)} & 97.2 {\std (0.9)} & 77.8 {\std (3.4)} & 70.0 {\std (4.0)} & 68.8 {\std (2.7)} & 57.0 {\std (2.8)} &  77.6 {\std 8.2} & 66.0 {\std (3.9)} & 8.5 \\
     
     GAE-2~\citep{kipf2016variational} & 58.3 {\std (5.1)} & 52.9 {\std (3.6)} & 64.0 {\std (3.1)} & 99.3 {\std (0.4)} & 85.4 {\std (4.0)} & 73.7 {\std (4.2)} & 68.2 {\std (3.3)} & 57.3 {\std (4.1)} & 75.5 {\std (14.7)} & 64.2 {\std (1.8)} & 8.9 \\
     
     GraphMAE-2~\citep{hou2022graphmae} & 61.5 {\std (3.3)} & 52.4 {\std (1.8)} &  62.1 {\std (4.3)} & 98.5 {\std (0.9)} & 83.0 {\std (3.4)} & 69.6 {\std (4.0)} &  67.2 {\std (2.7)} & 59.3 {\std (2.1)} &  77.1 {\std (14.2)} & 64.3 {\std (1.2)} & 8.3 \\
     
     \midrule 

     \method w/o $\mathbb{L}_{\mathbf{X}}$ &  80.4 {\std (2.2)} &  77.6 {\std (1.7)} &  70.1 {\std (0.8)} &  100.0 {\std (0.0)} & 83.6 {\std (2.5)} & 65.9 {\std (7.1)} & 66.0 {\std (3.3)} & 62.4 {\std (8.4)} & 67.3 {\std (15.2)} & 65.1 {\std (3.4)} & 5.4\\
     \method w/o $\mathbb{L}_{\mathbf{A}}$ & 56.9 {\std (6.9)} & 61.3 {\std (5.8)} & 63.2 {\std (1.2)} & 86.4 {\std (3.5)} & 73.4 {\std (4.6)} & 47.9 {\std (1.4)} & 56.3 {\std (4.6)} & 53.0 {\std (8.5)} & 65.8 {\std (12.9)} & 63.7 {\std (3.5)} & 12.9 \\
     \method w/o \texttt{AVG} & 77.9 {\std (2.8)} & 70.4 {\std (4.9)} & 70.1 {\std (0.7)} & 92.1 {\std (2.8)} & 82.5 {\std (4.2)} & 67.4 {\std (6.4)} & 63.2 {\std (3.4)} & 60.6 {\std (7.6)} & 59.6 {\std (11.5)} & 65.3 {\std (3.7)} & 8.4\\
     \method w/o \texttt{STD} & 74.8 {\std (4.0)} & 77.8 {\std (1.8)} & 62.9 {\std (2.9)} & 100.0 {\std (0.1)} & 84.0 {\std (4.0)} &  72.1 {\std (3.8)} & 64.5 {\std (1.8)} & 64.1 {\std (6.0)} & 64.3 {\std (10.2)} &  66.2 {\std (2.6)} & 4.9\\

     \midrule 
     
     \method &  81.0 {\std (2.0)} &  78.7 {\std (2.2)} &  72.5 {\std (0.3)} &  100.0 {\std (0.0)} &  84.7 {\std (4.4)} &  69.6 {\std (3.7)} &  68.1 {\std (3.0)} &  64.3 {\std (8.0)} & 68.2 {\std (11.3)}  &  67.6 {\std (4.0)}  & 2.7 \\
     
     \bottomrule
\end{tabular}
\label{tab:anomexp0}
}
\end{table}

\begin{table}[t]
\caption{Average and standard deviation of the test AUROC when graphs belonging to class 1 are anomalies and graphs belonging to class 0 are normal.
A.R. denotes average ranking.
\method outperforms other GLAD methods in terms of average ranking.
}
\setlength{\tabcolsep}{2pt}
\small
\centering
\scalebox{0.8}{
\renewcommand{\arraystretch}{1.1}
\begin{tabular}{l|c c c c c c c c c c c c c}
    \toprule
        Methods & DD & Protein & NCI1 & AIDS & Reddit & IMDB & MUTAG & DHFR & BZR & ER & A.R. \\
    \midrule 
    \midrule 
     
     DOMINANT~\citep{ding2019deep} & 84.0 {\std (5.8)} & 68.3 {\std (7.7)} & 64.8 {\std (6.1)} & 90.1 {\std (2.6)} & 82.3 {\std (5.1)} & 61.6 {\std (9.6)} & 70.2 {\std (2.6)} & 63.6 {\std (9.2)} &  80.0 {\std (5.3)} & 50.5 {\std (6.5)} & 6.7 \\
     
     OCGTL~\citep{qiu2022raising} & 71.0 {\std (7.6)} &  74.2 {\std (6.9)} & 58.2 \std{(7.0)} & 91.4 {\std (5.9)} & 62.6 {\std (4.4)} & 62.2 {\std (8.2)} & 66.6 {\std (3.1)} &  63.2 {\std (14.0)} & 67.3 {\std (22.9)} & 60.2 {\std (2.5)} & 9.5 \\
     
     GLocalKD~\citep{ma2022deep} & 25.9 {\std (5.6)} & 38.8 {\std (13.6)} & 57.2 {\std (4.5)} & 4.9 {\std (1.6)} & 20.5 {\std (5.5)} & 48.9 {\std (7.9)} & 54.3 {\std (6.1)} & 46.3 {\std (6.6)} &  48.8 {\std (20.6)} & 67.8 {\std (3.8)} & 15.8 \\
     
     GLADC~\citep{luo2022deep} & 29.5 {\std (6.7)} & 40.9 {\std (8.5)}  & 59.7 {\std (2.7)} &  4.7 {\std (1.6)} & 21.1 {\std (1.9)} & 46.7 {\std (6.9)} & 51.0 {\std (4.9)} & 49.6 {\std (5.3)} & 54.4 {\std (15.9)} & 65.7 {\std (2.1)} & 15.8 \\
     
     GLAM~\citep{zhao2022graph} & 74.0 {\std (5.3)} & 60.6 {\std (6.4)} & 55.4 {\std (1.9)} & 93.0 {\std (2.1)} & 78.9 {\std (3.5)} & 66.5 {\std (4.5)} & 65.3 {\std (2.4)} & 66.1 {\std (5.4)} & 83.0 {\std (9.5)} & 43.7 {\std (3.1)} & 8.1 \\
     
     HIMNET~\citep{niu2023graph} & 33.9 {\std (4.6)} & 47.8 {\std (8.6)} &  66.1 {\std (4.0)} &  30.6 {\std (5.9)}  & 58.8 {\std (2.9)} & 54.8 {\std (4.8)} & 56.4 {\std (3.3)} & 66.6 {\std (6.4)} & 74.4 {\std (12.7)} & 58.9 {\std (2.1)} & 11.5 \\
     
     SIGNET~\citep{liu2024towards} & 64.7 {\std (9.5)} & 54.6 {\std (6.8)} & 63.4 {\std (2.8)} &  96.2 {\std (1.0)} &  71.2 {\std (3.9)} & 42.1 {\std (5.6)} & 67.8 {\std (1.7)} & 62.2 {\std (4.5)} & 58.5 {\std (12.7)} & 55.8 {\std (5.3)} & 11.1 \\
     
     \midrule 

     
     GraphCL-1~\citep{you2020graph} & 79.9 {\std (2.2)} & 70.0 {\std (5.0)} & 45.2 {\std (2.3)} & 97.6 {\std (0.5)} & 56.6 {\std (5.7)} & 62.5 {\std (5.7)} & 53.6 {\std (3.1)} & 64.1 {\std (4.8)} & 64.4 {\std (10.7)} &  47.7 {\std (4.7)} & 10.7\\
     
     GAE-1~\citep{kipf2016variational} & 61.6 {\std (4.3)} & 45.8 {\std (9.1)} & 66.0 {\std (2.6)} & 72.2 {\std (2.7)} & 63.2 {\std (4.3)} & 66.9 {\std (6.6)} & 71.3 {\std (3.5)} & 54.0 {\std (7.6)} & 75.5 {\std (18.6)} & 59.8 {\std (4.2)} & 9.8 \\
     
     GraphMAE-1~\citep{hou2022graphmae} & 63.0 {\std (9.6)} & 58.5 {\std (6.3)} & 55.0 {\std (2.6)} & 93.3 {\std (1.0)} & 71.3 {\std (3.3)} & 62.1 {\std (4.0)} & 62.5 {\std (2.8)} & 72.1 {\std (9.9)} & 77.0 {\std (10.0)} & 39.0 {\std (2.6)} & 10.2 \\
     
     \midrule 
     
    GraphCL-2~\citep{you2020graph} & 73.2 {\std (2.7)} & 66.3 {\std (8.5)} & 58.3 {\std (4.5)} & 86.3 {\std (6.0)} & 76.8 {\std (4.7)} & 62.5 {\std (7.1)} & 66.0 {\std (3.8)} & 61.2 {\std (6.3)} &  66.1 {\std (12.5)} & 68.5 {\std (2.8)} & 9.3 \\
     
     GAE-2~\citep{kipf2016variational} & 76.0 {\std (1.6)} & 71.7 {\std (6.4)} & 60.8 {\std (4.6)} & 89.0 {\std (5.8)} & 65.2 {\std (7.3)} & 59.5 {\std (11.0)} & 66.3 {\std (3.2)} & 64.3 {\std (7.0)} & 68.5 {\std (2.9)} & 67.2 {\std (2.1)} & 8.3 \\
     
     GraphMAE-2~\citep{hou2022graphmae} & 74.4 {\std (5.3)} & 70.0 {\std (6.1)} &  74.4 {\std (2.8)} & 83.0 {\std (6.3)} & 68.5 {\std (6.1)} & 63.8 {\std (7.6)} &  69.0 {\std (2.1)} & 63.5 {\std (9.8)} &  68.5 {\std (6.4)} & 68.1 {\std (3.0)} & 6.7 \\
     
     \midrule 

     \method w/o $\mathbb{L}_{\mathbf{X}}$ &  78.4 {\std (5.2)} &  73.5 {\std (5.7)} &  68.3 {\std (3.8)} &  99.3 {\std (0.9)} & 60.8 {\std (5.4)} & 65.6 {\std (4.3)} & 65.5 {\std (2.8)} & 58.3 {\std (4.8)} & 63.9 {\std (23.6)} & 67.4 {\std (3.7)} & 7.5 \\
     \method w/o $\mathbb{L}_{\mathbf{A}}$ & 66.6 {\std (8.3)} & 68.1 {\std (8.3)} & 63.0 {\std (5.3)} & 92.1 {\std (2.1)} & 69.9 {\std (4.9)} & 65.9 {\std (12.8)} & 57.7 {\std (2.3)} & 63.1 {\std (8.5)} & 71.5 {\std (15.4)} & 57.7 {\std (4.5)} & 9.7 \\
     \method w/o \texttt{AVG} & 79.3 {\std (5.1)} & 65.7 {\std (6.0)} & 65.8 {\std (3.3)} & 97.8 {\std (2.3)} & 63.9 {\std (8.9)} & 64.9 {\std (6.6)} & 58.6 {\std (4.4)} & 59.6 {\std (12.5)} & 73.0 {\std (14.4)} & 58.6 {\std (3.3)} & 8.8\\
     \method w/o \texttt{STD} & 73.7 {\std (6.8)} & 71.0 {\std (8.6)} & 67.4 {\std (4.2)} & 97.3 {\std (0.9)} & 57.0 {\std (4.5)} &  69.3 {\std (3.5)} & 59.5 {\std (3.0)} & 61.7 {\std (6.7)} & 78.2 {\std (12.7)} &  67.1 {\std (2.2)} & 7.3 \\

     \midrule 
     
     \method &  79.9 {\std (3.5)} &  78.0 {\std (2.1)} &  69.6 {\std (3.7)} &  99.3 {\std (1.0)} &  72.1 {\std (6.9)} &  68.7 {\std (3.3)} &  66.9 {\std (3.7)} &  63.2 {\std (9.2)} & 70.8 {\std (13.9)}  &  68.2 {\std (3.2)} & 3.6  \\
     
     \bottomrule
\end{tabular}
\label{tab:anomexp1}
}
\vspace{5mm}
\end{table}

We present experimental results in each setting of our main experiment. 
Specifically, experimental results for the case where class 0 graphs are anomalies and class 1 graphs are normal are provided in Table~\ref{tab:anomexp0}. 
In addition, experimental results for the case where class 1 graphs are anomalies and class 0 graphs are normal are provided in Table~\ref{tab:anomexp1}.
Notably, in both cases, \method obtains the best average ranking among 18 methods (refer to Table~\ref{tab:anomexp0} and Table~\ref{tab:anomexp1}).

\subsection{Experimental results on additional evaluation metrics}

\begin{table}[t]
\caption{\textbf{GLAD performance}: 
Average and standard deviation of test \textbf{AP score} values ($\times$100) in the GLAD task are reported.
The \textcolor{sunwoogreen2}{best} performances are highlighted in \textcolor{sunwoogreen2}{green}.}
\setlength{\tabcolsep}{2.5pt}
\small
\centering
\scalebox{0.92}{
\renewcommand{\arraystretch}{1.0}
\begin{tabular}{l|c c c c c c c c c c}
    \toprule
     Method & DD & Protein & NCI1 & AIDS & Reddit & IMDB & MUTAG & DHFR & BZR & ER \\
    \midrule 
    \midrule 
     
     OCGTL~\citep{qiu2022raising} & 84.9 {\std (3.1)} &  78.1 {\std (1.8)} & 73.6 \std{(1.6)} & 95.3 {\std (3.0)} & \best 88.0 {\std (1.8)} & 81.0 {\std (2.2)} & 72.7 {\std (1.9)} & \best 87.6 {\std (4.1)} & \best 88.6 {\std (0.9)} & 60.3 {\std (1.1)} \\
     
     GLAM~\citep{zhao2022graph} & 74.7 {\std (1.7)} & 71.1 {\std (2.0)} & 73.6 {\std (9.0)} & 95.4 {\std (2.4)} & 83.2 {\std (1.6)} & 78.7 {\std (3.0)} & 75.9 {\std (1.5)} & 75.4 {\std (3.2)} & 76.9 {\std (1.2)} & 65.0 {\std (0.6)} \\
     
     \midrule 
     
     \method & \best 88.1 {\std (1.3)} & \best 86.5 {\std (1.3)} & \best 81.8 {\std (1.2)} & \best 99.7 {\std (0.6)} & 81.7 {\std (2.0)} & \best 81.7 {\std (2.7)} & \best 79.6 {\std (1.4)} & 78.9 {\std (2.8)} & 79.1 {\std (2.1)}  & \best 70.1 {\std (0.4)} \\
     
     \bottomrule
\end{tabular}
\label{tab:apscore}
}
\end{table}

\begin{table}[t]
\caption{\textbf{GLAD performance}: 
Average and standard deviation of test \textbf{Precision@10 score} values ($\times$10) in the GLAD task are reported.
The \textcolor{sunwoogreen2}{best} performances are highlighted in \textcolor{sunwoogreen2}{green}.}
\setlength{\tabcolsep}{2.5pt}
\small
\centering
\scalebox{1.0}{
\renewcommand{\arraystretch}{1.0}
\begin{tabular}{l|c c c c c c c c c c}
    \toprule
     Method & DD & Protein & NCI1 & AIDS & Reddit & IMDB & MUTAG & DHFR & BZR & ER \\
    \midrule 
    \midrule 
     
     OCGTL~\citep{qiu2022raising} & 5.2 {\std (0.7)} &  3.5 {\std (0.7)} & 4.4 \std{(0.8)} & 9.8 {\std (0.2)} & 3.5 {\std (0.6)} & 3.9 {\std (0.6)} & 5.1 {\std (1.2)} & 3.4 {\std (0.9)} & 4.0 {\std (0.6)} & 5.1 {\std (0.4)} \\
     
     GLAM~\citep{zhao2022graph} & 5.1 {\std (0.9)} & 6.0 {\std (0.7)} & 5.1 {\std (1.0)} & 9.9 {\std (0.4)} & \best 8.4 {\std (0.7)} & 5.9 {\std (0.8)} & 4.9 {\std (0.8)} & 4.1 {\std (0.6)} & \best 5.5 {\std (0.6)} & 4.7 {\std (0.5)} \\
     
     \midrule 
     
     \method & \best 7.7 {\std (1.3)} & \best 7.9 {\std (1.3)} & \best 7.8 {\std (1.3)} & \best 10.0 {\std (0.0)} & 5.9 {\std (1.7)} & \best 6.6 {\std (1.6)} & \best 8.5 {\std (1.1)} & \best 5.2 {\std (1.1)} & 4.8 {\std (1.3)}  & \best 5.3 {\std (0.9)} \\
     
     \bottomrule
\end{tabular}
\label{tab:precision}
}
\end{table}

\begin{table}[t]
\caption{\textbf{Additional ablation study}: 
Average and standard deviation of test AUROC values ($\times$10) in the GLAD task are reported.
The \textcolor{sunwoogreen2}{best} performances are highlighted in \textcolor{sunwoogreen2}{green}.}
\setlength{\tabcolsep}{2.5pt}
\small
\centering
\scalebox{0.9}{
\renewcommand{\arraystretch}{1.0}
\begin{tabular}{l|c c c c c c c c c c}
    \toprule
     Method & DD & Protein & NCI1 & AIDS & Reddit & IMDB & MUTAG & DHFR & BZR & ER \\
    \midrule 
    \midrule 
     
    \method w/o Aug. & 78.9 {\std (3.0)} & 77.1 {\std (3.2)} & 68.8 \std{(2.3)} & 99.6 {\std (0.1)} & 74.2 {\std (3.3)} & 69.1 {\std (3.6)} & 66.0 {\std (2.3)} & \best 64.4 {\std (5.4)} & 66.4 {\std (12.9)} & \best 69.9 {\std (3.0)} \\
     
     \method w/o Cos. & 79.5 {\std (2.4)} & 72.8 {\std (3.4)} & 68.4 {\std (3.7)} & 96.5 {\std (2.8)} &  72.7 {\std (6.7)} & 63.3 {\std (3.8)} & 65.1 {\std (3.1)} & 60.1 {\std (4.1)} &  59.9 {\std (9.7)} & 68.3 {\std (2.5)} \\
     
     \midrule 
     
     \method & \best 80.5 {\std (2.3)} & \best 78.4 {\std (2.2)} & \best 71.1 {\std (2.0)} & \best 99.7 {\std (0.5)} & \best  78.4 {\std (5.7)} & \best 69.2 {\std (3.5)} & \best 67.5 {\std (3.4)} &  63.8 {\std (8.6)} & \best 67.5 {\std (12.6)} &  67.9 {\std (3.6)} \\
     
     \bottomrule
\end{tabular}
\label{tab:abltable}
}
\end{table}

\begin{table}[t]
\caption{\textbf{Large-scale graphs GLAD performance}: 
The average and standard deviation of test \textbf{AUROC} values ($\times$10) in the GLAD task on large-scale graphs are reported.
The \textcolor{sunwoogreen2}{best} performances are highlighted in \textcolor{sunwoogreen2}{green}.}
\small
\centering
\scalebox{1.0}{
\renewcommand{\arraystretch}{1.0}
\begin{tabular}{l|c c c c c c c c c c}
    \toprule
     Method & OCGTL~\citep{qiu2022raising} & GLAM~\citep{zhao2022graph} & \method-Sample & \method \\
    \midrule 
    \midrule 
     
     MalNetTiny & 60.5 {\std (4.3)} & 56.8 {\std (4.3)} & 66.0 {\std (2.5)} &  68.1 {\std (2.7)} \\ 
     OVCAR-8 & 69.8 {\std (7.7)} & 66.0 {\std (3.5)} & 72.2 {\std (8.6)} &  72.8 {\std (8.2)} \\ 
     
     \bottomrule
\end{tabular}
\label{tab:large_scale}
}
\end{table}

\begin{table}[t]
\caption{\textbf{Runtime analysis:} 
Average runtime (secs) per each graph of each method on each dataset.
Overall, \method is the second-fasted method among the four methods.}
\small
\centering
\scalebox{0.75}{
\renewcommand{\arraystretch}{1.0}
\begin{tabular}{l|c c c c c c c c c c c c }
    \toprule
     Method & DD & Protein & NCI1 & AIDS & Reddit & IMDB & MUTAG & DHFR & BZR & ER & MalNetTiny & OVCAR-8\\
    \midrule 
    \midrule 
     GLAM~\citep{zhao2022graph} & 0.003 & 0.002 & 0.002 & 0.002 & 0.005 & 0.002 & 0.002 & 0.003 & 0.002 & 0.002 & 0.01 & 0.003 \\
     OCGTL~\citep{qiu2022raising} & 0.005 & 0.004 & 0.004 & 0.004 & 0.007 & 0.004 & 0.004 & 0.004 & 0.004 & 0.004 & 0.01 & 0.006 \\
     \method-Sample & 0.006 & 0.004 & 0.004 & 0.004 & 0.012 & 0.004 & 0.004 & 0.004 & 0.004 & 0.004 & 0.05 & 0.004 \\
     \method & 0.005 & 0.003 & 0.003 & 0.003 & 0.007 & 0.003 & 0.003 & 0.004 & 0.003 & 0.003 & 0.03 & 0.004 \\
     \bottomrule
\end{tabular}
\label{tab:runtime}
}
\end{table}

In this section, we compare the GLAD performance of \method against those of two strongest competitors, OCGTL and GLAM. 
To this end, we leverage two metrics: average precision score (AP score) and precision@K.
As shown in Table~\ref{tab:apscore} and Table~\ref{tab:precision}, \method outperforms the competitors in 7/10 and 8/10 in terms of AP score and precision@K, respectively.
These results demonstrate the effectiveness of \method is not limited to a particular evaluation metric.

\subsection{Experimental results on additional ablation study}

In this section, we demonstrate the effectiveness of the following two components of \method: graph augmentation and cosine-similarity feature reconstruction loss.
To this end, we use two variants of \method.
\textbf{\method w/o Aug.} is a variant where the input graph augmentation is removed.
\textbf{\method w/o Cos.} is a variant that uses the Frobenious norm loss for the feature reconstruction loss, instead of the cosine similarity loss. 
Note that \method uses both input graph augmentation and cosine similarity loss.
As shown in Table~\ref{tab:abltable}, \method outperforms its variants in 8 out of 10 datasets, demonstrating the effectiveness of our graph augmentation and cosine-similarity loss.

\subsection{Experimental results from the scalability analysis}\label{subsec:scalability}

In this section, we study the scalability and effectiveness of GLAD methods in large-scale graphs.

\textbf{Datasets.}
We use two large-scale real-world graphs, which are MalNetTiny~\citep{freitas2020large} and OVCAR-8~\cite{yan2008mining}. 
Specifically, the MalNetTiny dataset consists of graphs, each containing a large number of nodes and edges.
In contrast, the OVCAR-8 dataset consists of a large number of graphs.

\textbf{Scalable \method.}
Note that \method may not be scalable for graphs with a large number of nodes, since \method reconstructs all the entries of an adjacency matrix, which results in the complexity of $O(\vert \mathcal{V}\vert^{2})$.
Thus, we present a scalable version of \method, which we call \textbf{\method-Sample}. 
Specifically, \method-Sample samples $K$ number of entries from the adjacency matrix, and reconstructs only the sampled entries.
Therefore, the time complexity of computing reconstruction loss becomes
\begin{equation}
    {O(\vert \mathcal{V} \vert K )} \equiv {O(\vert \mathcal{V} \vert)}, \ \because \text{$K$ is a constant, which is the same across all graphs.}
\end{equation}

\textbf{Experimental results.}
We compare the (1) GLAD performance and (2) inference runtime of our proposed methods (i.e., \method and \method-Sample) against the two strongest baseline methods: OCGTL and GLAM.
Regarding the performance comparison, as shown in Table~\ref{tab:large_scale}, 
\method and \method-Sample outperform the baseline methods in both datasets.
Regarding the runtime analysis, as shown in Table~\ref{tab:runtime}, 
\method is the second-fastest method among the four methods.



\section{Further Analysis and Discussion}
\label{sec:intuitiondiscussion}
In this section, we provide our further analyses and discussions.

\subsection{Reconstruction flip in computer vision}
\label{subsec:cvreconflip}

Analogous to our observations of the reconstruction flips in graphs, certain images are more easily reconstructed. 
For instance, \citet{liu2023diversity} noted that "\textit{anomalies with colors close to the background may yield unreliable reconstruction errors}".
Moreover, in the MNIST dataset, a reconstruction model trained on class-7 images sometimes reconstructs class-4 images better than class-7 images~\citep{liu2023diversity}.

To mitigate this undesirable phenomenon, \citet{liu2023diversity} proposed a deformation-based anomaly detection method.
In this method, an input image is first deformed in a way that makes a reconstruction model easy to reconstruct.
Here, a deformation method that maintains the semantics of the input image is employed~\citep{dai2017deformable}.
To identify whether an image is an anomaly, they first deform the image and then let a reconstruction model reconstruct it.
Their underlying intuition is that the deformation model would drastically change the image dissimilar to the majority of the training images, and this would make the deformed anomalous image different from its original image. 
Finally, when the deformed image is reconstructed by the reconstruction model, the reconstructed anomalous image would be significantly different from the original anomalous image, resulting in a large reconstruction error.

While the method is intuitive in computer vision, the extension of the suggested method to the graph domain is non-trivial.
First, the graph deformation that maintains semantics is non-trivial.
It is hard to know the semantics of a graph without sufficient domain knowledge, especially without any external graph labels.
Furthermore, naive graph augmentations, such as node dropping and edge perturbation, may harm the semantics of a graph~\citep{xia2022simgrace}.
Second, while the deformation techniques are widely studied in computer vision~\citep{dai2017deformable, chen2019image}, those of a graph is an underexplored region.
Therefore, employing an adequate technique for graph deformation is also practically infeasible.

Considering these challenges, we consider that the method proposed in~\citet{liu2023diversity} is hard to be trivially extended to the graph domain.

\subsection{Reconstruction flip in real-world graph datasets}
\label{subsec:realworldrecon}

We provide our in-depth analysis result of reconstruction flip in Section~\ref{sec:analysis} by using the \textit{synthetic datasets}.
In this section, we analyze whether the reconstruction flip phenomena occur in \textit{real-world graph datasets}. 

\textbf{Setting.} 
We focus on the anomaly detection task performance of GAE~\cite{kipf2016variational}, which reconstructs the adjacency matrix of a given graph.
GAE uses reconstruction loss as an anomaly score, where an anomalous graph is likely to have high reconstruction errors. 
We measure the AUROC score on test graphs, based on the setting described in Section~\ref{subsec:expsetting}.
Here, the AUROC score lower than 0.5 indicates the corresponding model performs worse than random guessing, implying the reconstruction flip has occurred.

\textbf{Results.} As shown in Table~\ref{tab:anomexp0}, in the protein dataset, GAE~\citep{kipf2016variational} tends to perform worse than random guessing (spec., average AUROC is 0.458), which indicates that the reconstruction flip occurs.
Thus, we demonstrate that the reconstruction flip is not limited to synthetic scenarios but also occurs in real-world graph datasets.

\subsection{Complexity analysis of \method}\label{subsec:timecomplexity}

In this section, we provide a time complexity analysis for the forward pass of \method, concerning the size of a graph (i.e., number of nodes $\vert \mathcal{V}\vert$ and edges $\vert \mathcal{E}\vert$).

\textbf{Encoding.}
We first encode a graph $\mathcal{G}= (\mathbf{X} \in \mathbb{R}^{\vert \mathcal{V}\vert \times d},\mathbf{A} \in \{0,1\}^{\vert \mathcal{V}\vert \times \vert \mathcal{V}\vert})$ with a GNN encoder.
The time complexity of encoding $\mathcal{G}$ with $L-$layer GCN~\citep{kipf2016semi} is ${O}(L\vert \mathcal{V} \vert d^{2} + L \vert \mathcal{E}\vert d)$~\citep{chiang2019cluster}.
Given that $d$ and $L$ are constants, the complexity is equivalent to $\mathcal{O}(\vert \mathcal{V} \vert + \vert \mathcal{E} \vert)$.~\footnote{Note that the forward pass time complexity of GIN~\citep{xu2018powerful} is equivalent to that of GCN concerning graph size. This is because the difference between GCN and GIN lies in the feature transformation, which is independent of the graph size.}

\textbf{Decoding.}
We decode node embeddings by using a node decoder and edge decoder, which are MLPs.
Since the hidden dimensions and the number of layers of MLPs are all constants, the complexity of decoding is equivalent to $O(\vert \mathcal{V}\vert)$.

\textbf{Reconstruction loss.}
We compute adjacency matrix reconstruction loss and node feature reconstruction loss. 
Each loss computation has the time complexity of $O({\vert \mathcal{V}\vert^{2}})$ and $O(\mathcal{\vert \mathcal{V}\vert})$, respectively. 
Thus, the overall time complexity of the reconstruction loss computation becomes $O(\vert \mathcal{V}\vert^{2})$.

\textbf{Error representation.}
We utilize average pooling and standard deviation pooling, which both have linear time complexity.
Thus, the time complexity is equivalent to $O({\vert \mathcal{V}\vert^{2}})$.

\textbf{Overall.}
Thus, the time complexity of the forward pass of \method is as follows:
\begin{equation}
    O(\vert \mathcal{V}\vert + \vert \mathcal{E}\vert) + O(\vert \mathcal{V}\vert) + O(\vert \mathcal{V}\vert^{2}) + O(\vert \mathcal{V}\vert^{2}) = O(\vert \mathcal{V}\vert^{2}).
\end{equation}
While \method has a quadratic complexity to the number of nodes, our task is a graph-level task where real-world graphs typically have an affordable number of nodes (as shown in Table~\ref{tab:datasets}, the dataset that has the largest average number of nodes has smaller than $430$ average number of nodes).





\subsection{Broader impacts of our work}\label{subsec:broaderimpact}

In this section, we discuss the broader impacts of our research.
While our focus lies in the graph-level task, similar observations can be found in node-level tasks within various types of graphs, such as dynamic graphs~\cite{lee2024slade}, heterophilic graphs~\cite{lee2024feature, liang2024sign}, and hypergraphs~\cite{kim2024hypeboy, kim2024survey, choe2023classification, kim2023datasets}.
We anticipate that our findings and the proposed \method will be widely utilized in applications requiring graph-level anomaly detection, such as drug discovery and brain diagnosis.
In addition, as described in Section~\ref{subsec:relatedwork}, the phenomenon of the reconstruction flip is not limited to the graph domain, and it is also observed in other domains such as computer vision. 
To our knowledge, we are the first to utilize summarized reconstruction errors as a representation feature of data. 
This approach can be directly applied to other fields, such as computer vision.

\subsection{Generalized categorization of primary patterns}

\textbf{Expectation.} In this work, as a primary pattern, we use community structure and node cycles.
As a future direction, more general and formal categorization of graph patterns can be considered.
This categorization would enable our analysis (Section~\ref{sec:analysis}) to be more systematic and generalized.

\textbf{Challenge.}
However, real-world datasets may exhibit a wider variety of patterns, including those related to node attributes.
For example, in the protein dataset, both normal and anomalous graphs exhibit homophilic patterns (i.e., edges tend to join nodes with the same attribute), with anomalies displaying a stronger pattern than normal graphs.
This diversity makes the categorization not straightforward.
Thus, adequately considering real-world graph patterns would play a crucial role in this categorization.



\newpage

\section{Checklist}


\section*{NeurIPS Paper Checklist}

\begin{enumerate}

\item {\bf Claims}
    \item[] Question: Do the main claims made in the abstract and introduction accurately reflect the paper's contributions and scope?
    \item[] Answer: \answerYes{}
    \item[] Justification: We clarify our goal, analyzing reconstruction flip, and provide analysis results in Section~\ref{sec:analysis}. 
    We present a method (Section~\ref{sec:method}) and demonstrate its superiority (Section~\ref{sec:experiment}).

\item {\bf Limitations}
    \item[] Question: Does the paper discuss the limitations of the work performed by the authors?
    \item[] Answer: \answerYes{}
    \item[] Justification: We provide limitations of the proposed method \method in Section~\ref{sec:conclusion}.

\item {\bf Theory Assumptions and Proofs}
    \item[] Question: For each theoretical result, does the paper provide the full set of assumptions and a complete (and correct) proof?
    \item[] Answer: \answerYes{}
    \item[] Justification: We provide the full statements for all theorems in Section~\ref{subsec:theoreticalanalysis} and provide full proofs in Appendix~\ref{sec:appendix_proof}.

    \item {\bf Experimental Result Reproducibility}
    \item[] Question: Does the paper fully disclose all the information needed to reproduce the main experimental results of the paper to the extent that it affects the main claims and/or conclusions of the paper (regardless of whether the code and data are provided or not)?
    \item[] Answer: \answerYes{}
    \item[] Justification: We provide architectures of \method in Section~\ref{sec:method}, details regarding hyperparameters in Appendix~\ref{sec:experimentdetails}, and model code in \url{https://github.com/kswoo97/GLAD_MUSE}.

\item {\bf Open access to data and code}
    \item[] Question: Does the paper provide open access to the data and code, with sufficient instructions to faithfully reproduce the main experimental results, as described in supplemental material?
    \item[] Answer: \answerYes{}
    \item[] Justification: We provide all the utilized datasets, full code for models, and instructions to run the code in \url{https://github.com/kswoo97/GLAD_MUSE}.

\item {\bf Experimental Setting/Details}
    \item[] Question: Does the paper specify all the training and test details (e.g., data splits, hyperparameters, how they were chosen, type of optimizer, etc.) necessary to understand the results?
    \item[] Answer: \answerYes{}
    \item[] Justification: We provide the key setting in Section~\ref{sec:experiment}, and further details in Appendix~\ref{sec:experimentdetails}.

\item {\bf Experiment Statistical Significance}
    \item[] Question: Does the paper report error bars suitably and correctly defined or other appropriate information about the statistical significance of the experiments?
    \item[] Answer: \answerYes{}
    \item[] Justification: We describe the number of experimental runs and standard deviation of the performance in Section~\ref{sec:experiment}.

\item {\bf Experiments Compute Resources}
    \item[] Question: For each experiment, does the paper provide sufficient information on the computer resources (type of compute workers, memory, time of execution) needed to reproduce the experiments?
    \item[] Answer: \answerYes{}
    \item[] Justification: We describe machines utilized for the experiments in Appendix~\ref{subsec:machines}.
    
\item {\bf Code Of Ethics}
    \item[] Question: Does the research conducted in the paper conform, in every respect, with the NeurIPS Code of Ethics \url{https://neurips.cc/public/EthicsGuidelines}?
    \item[] Answer: \answerYes{}
    \item[] Justification: We strictly follow the NeurIPS Code of Ethics, particularly in terms of anonymity that we do not reveal our nationality and institutes, etc.

\item {\bf Broader Impacts}
    \item[] Question: Does the paper discuss both potential positive societal impacts and negative societal impacts of the work performed?
    \item[] Answer: \answerYes{}
    \item[] Justification: We discuss the broader impact of our work in Appendix~\ref{subsec:broaderimpact}

\item {\bf Safeguards}
    \item[] Question: Does the paper describe safeguards that have been put in place for responsible release of data or models that have a high risk for misuse (e.g., pretrained language models, image generators, or scraped datasets)?
    \item[] Answer: \answerNA{}
    \item[] Justification: We believe we do not have risks of misusing.

\item {\bf Licenses for existing assets}
    \item[] Question: Are the creators or original owners of assets (e.g., code, data, models), used in the paper, properly credited and are the license and terms of use explicitly mentioned and properly respected?
    \item[] Answer: \answerYes{}
    \item[] Justification: We properly add citations for all the mentioned existing literature or methods.

\item {\bf New Assets}
    \item[] Question: Are new assets introduced in the paper well documented and is the documentation provided alongside the assets?
    \item[] Answer: \answerNA{}
    \item[] Justification: We do not release any new assets.

\item {\bf Crowdsourcing and Research with Human Subjects}
    \item[] Question: For crowdsourcing experiments and research with human subjects, does the paper include the full text of instructions given to participants and screenshots, if applicable, as well as details about compensation (if any)? 
    \item[] Answer: \answerNA{}
    \item[] Justification: We do not utilize any crowdsourcing or human-related tasks.

\item {\bf Institutional Review Board (IRB) Approvals or Equivalent for Research with Human Subjects}
    \item[] Question: Does the paper describe potential risks incurred by study participants, whether such risks were disclosed to the subjects, and whether Institutional Review Board (IRB) approvals (or an equivalent approval/review based on the requirements of your country or institution) were obtained?
    \item[] Answer: \answerNA{}.
    \item[] Justification: We do not utilize any crowdsourcing or human-related tasks.

\end{enumerate}
\label{sec:checklist}



\end{document}